\ifdefined\pdfminorversion
\fi

\PassOptionsToPackage{table}{xcolor}
\documentclass{fairmeta}

\usepackage{amsmath,amsfonts,bm}

\def\1{\bm{1}}

\DeclareMathAlphabet{\mathsfit}{\encodingdefault}{\sfdefault}{m}{sl}
\SetMathAlphabet{\mathsfit}{bold}{\encodingdefault}{\sfdefault}{bx}{n}

\newcommand{\R}{\mathbb{R}}

\usepackage{hyperref}
\usepackage{url}
\usepackage{booktabs}
\usepackage{multirow}
\usepackage{mathtools}
\usepackage{algorithm}
\usepackage{algpseudocode}
\usepackage{listings}
\usepackage{amssymb}
\ifPDFTeX
  \usepackage[accsupp]{axessibility}
\fi 
\usepackage[table]{xcolor}

\definecolor{lightgreen}{RGB}{220,245,220}
\definecolor{lightblue}{RGB}{220,235,250}
\definecolor{lightyellow}{RGB}{250,245,210}
\definecolor{lightgray}{RGB}{242,242,242}
\definecolor{MyDarkGreen}{RGB}{0, 100, 0} 
\definecolor{darkgreen}{RGB}{35,115,65}
\definecolor{darkblue}{RGB}{45,90,155}
\definecolor{darkyellow}{RGB}{155,115,0}

\newcommand{\greenhl}[1]{\cellcolor{lightgreen}#1}
\newcommand{\bluehl}[1]{\cellcolor{lightblue}#1}

\usepackage{xspace}
\usepackage{comment}
\usepackage{enumitem}
\newcommand{\method}{\text{MaLoW}\xspace}

\title{Memory as Weights: Internalizing Long-Term History for Streaming Videos}

\author[1]{\href{https://scholar.google.com/citations?user=HXOVwRAAAAAJ&hl=en}{Zhuoming Liu}}
\author[1]{\href{https://scholar.google.com/citations?user=v4UwIYMAAAAJ&hl=en}{Jinhong Lin}}
\author[1]{\href{https://scholar.google.com/citations?user=5-PDQ94AAAAJ&hl=en}{Lin Zhang}}
\author[2]{\href{https://scholar.google.com/citations?user=irrbH_IAAAAJ&hl=en}{Yiwu Zhong}}
\author[1]{\href{https://scholar.google.com/citations?user=_y-8nrcAAAAJ&hl=en}{Yin Li}}
\affiliation[1]{University of Wisconsin--Madison}
\affiliation[2]{Peking University}

\abstract{%
  \begingroup
  \renewenvironment{abstract}{}{}%
  \begin{abstract}
Streaming vision-language models (VLMs) must continuously integrate information from the past while remaining responsive to current observations and user queries. Existing memory mechanisms typically compress, retrieve, or cache historical observations and reintroduce them into the input context, forcing historical memory and current observations to compete for limited context. We propose \textit{\method}, a streaming memory framework that instead internalizes historical context into evolving lightweight parameter updates. \method uses a recurrent HyperNetwork to incrementally consolidate incoming video observations into fixed-size LoRA weights, allowing historical information to be carried forward without revisiting the full video stream. In doing so, \method represents \textit{history as weights, while keeping the current observation as input}. To support training, we further construct \textit{\method-Instruct}, a large-scale streaming video dataset containing 381K question-answer pairs over 56K videos. We conduct extensive experiments to demonstrate that \method achieves strong performance on OVO-Bench, StreamingBench, and StreamArena. 
On the backward split of OVO-Bench, \method outperforms prior memory approaches by 3.9\%--11.8\% across different video backbones while incurring only modest latency overhead. 
Beyond streaming video understanding, \method also generalizes to vision-language-action models, improving the action success rate on Simpler WidowX by 9.3\%.

\end{abstract}%
  \endgroup
}

\metadata[Project page]{\url{https://dragonlzm.github.io/MaLoW}}
\metadata[Code]{\url{https://github.com/dragonlzm/MaLoW}}
\metadata[Models \& data]{\url{https://huggingface.co/collections/zhuomingliu/malow}}

\begin{document}
\maketitle
\section{Introduction}

By continuously processing an input video stream and responding to user queries, a streaming vision-language model (VLM) must reason over what has already happened, interpret what is happening now, and remain attentive to what may happen next. Such capabilities are central to key tasks ranging from streaming video understanding~\citep{niu2025ovo} to interactive AI assistants~\citep{thinkingmachines2026interactionmodels} and robotic planning~\citep{li24simpler}, where responses and decisions must be made online under latency and memory constraints. Streaming VLMs have thus emerged as an important research frontier. Recent work has shown that offline VLMs can be adapted to describe ongoing activities in streaming videos~\citep{shen2026simple, lu2026aura}, while instruction fine-tuning can further enable models to monitor future events~\citep{li2026videochat3, wang2026moss}.

\begin{figure*}[t]
    \centering
    \includegraphics[width=0.9\linewidth]{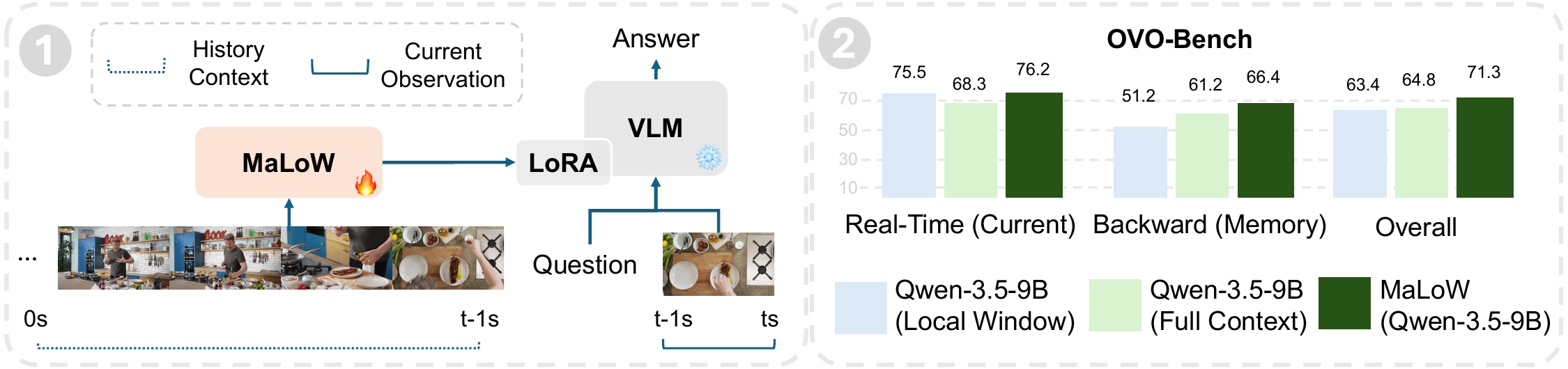}
    \vspace{-0.5em}
    \caption{\textbf{(1)}: \textbf{\method} internalizes the history context into time-evolving, lightweight parameter updates (LoRA) of a streaming VLM. At each step, the streaming VLM takes the current observation and user query as input, and performs inference using the updated LoRA weights. \textbf{(2)}: By separating memory from the input, \method preserves the performance on real-time perception while significantly improves the performance of memory-critical, backward tracing tasks.}
    \label{fig:teaser}
    \vspace{-1.4em}
\end{figure*}

Despite these efforts, \textit{a fundamental open problem remains: how should a streaming VLM remember?} As in the offline setting, an effective memory mechanism must preserve important visual information within a bounded memory footprint. The streaming setting, however, introduces additional challenges. \textit{First}, memory must evolve incrementally as new observations arrive, rather than being repeatedly recomputed from the full history. \textit{Second}, historical information must be incorporated without crowding out the current observation (see Fig. \ref{fig:teaser}-2). 
\textit{Finally}, a practical memory mechanism should introduce minimal inference overhead. 

To address these challenges, we propose \textbf{\method} (\underline{M}emory \underline{a}s \underline{Lo}RA \underline{W}eights), a framework that internalizes historical context into evolving lightweight parameter updates of a streaming VLM, as illustrated in Fig.~\ref{fig:teaser}-1. \method incrementally consolidates incoming observations into fixed-size LoRA weights, allowing accumulated history to be carried forward without revisiting the full video stream. The base VLM architecture and parameters stay unchanged and only the lightweight LoRA memory evolves over time. More fundamentally, rather than placing memory and current observations in competition within the same context window, as in prior work~\citep{guan2026video, liu2026thinking, zhang2026hermes}, \textit{\method represents history as weights while keeping the current observation as input}. At each step, the model takes the current observation and user query as input and performs inference using the updated LoRA weights that encode historical information. By separating historical memory from the input context, \method preserves the model's capacity to attend to current observations while retaining information from the past (Fig.~\ref{fig:teaser}-2), with only modest inference overhead.

Specifically, \method employs a recurrent HyperNetwork~\citep{hypernetwork} to efficiently generate the LoRA weights that encode historical information. At each time step, the incoming video clip and its textual context are first compressed into a compact representation, which is then used to update a set of recurrent hidden states. To model the temporal dynamics of these hidden states, \method maintains two parallel branches: a state-space model~\citep{yang2024parallelizing} for modeling local temporal dependencies, and a gated summation branch for preserving global historical semantics. The resulting states are mapped through a linear projection to the LoRA weights used by the VLM. To facilitate training, we further construct \textbf{\method-Instruct}, a comprehensive streaming video dataset containing 381K question-answer (QA) pairs over 56K videos, by adapting public offline video understanding datasets to the streaming setting.

To evaluate our framework, we conduct extensive experiments across benchmarks and settings. \method demonstrates state-of-the-art results on streaming video benchmarks, including OVO-Bench~\citep{niu2025ovo}, StreamingBench~\citep{lin2026streamingbench} and StreamArena~\citep{zhang2026streamarena}. 
On the backward split of OVO-Bench, for example, \method outperforms prior memory approaches by 3.9\%--11.8\% across different video backbones, while incurring modest latency overhead. 
Across base VLMs ranging from 0.8B to 11B parameters, \method yields consistent performance gains. Beyond streaming video understanding, we further extend \method to the Vision-Language-Action (VLA) setting, where incorporating historical observations and action trajectories improves the action success rate on Simpler~\citep{li24simpler} WidowX by 9.3\%. 

Our \textbf{contributions} are summarized as follows:
\begin{itemize}[leftmargin=2em, itemsep=2pt, topsep=2pt]
    \item We propose \method, a new \textit{streaming memory framework that internalizes historical context} into evolving LoRA weights, introducing a distinct design in which history is represented as weights while current observations remain as input.
    \item We develop a recurrent HyperNetwork that incrementally updates LoRA memory from streaming video and other context. We also construct \method-Instruct, a large-scale training dataset for streaming video understanding, which \textit{we will release publicly}.
    \item We demonstrate that \method achieves competitive results across streaming video benchmarks, attains consistent gains over diverse base VLMs, and generalizes to VLA settings.
\end{itemize}

\section{Related Work}

\noindent\textbf{Vision-Language Models.}
Vision-language models (VLMs)~\citep{liu2023visual, alayrac2022flamingo, zhang2024llava, clark2026molmo2, qwen3.5} unify diverse visual understanding and reasoning tasks through a natural-language interface. Among these tasks, long-video understanding~\citep{zhou2025mlvu, wang2025lvbench} is particularly challenging because models must preserve and reason over information spanning extended temporal contexts. 

Prior works~\citep{NEURIPS2024_d7ce06e9, he2024ma, song2024moviechat} have explored various memory and context-management mechanisms for this offline long video understanding setting. 
Our work builds on pretrained VLMs and equips them with an evolving long-term memory that is needed in the streaming video setting and incrementally updated as new observations arrive.

\medskip
\noindent\textbf{Streaming VLMs with Memory.}
Recent works on streaming VLMs~\citep{xia2026streaming, chen2024videollm, li2026videochat3, wang2026moss} explore several complementary directions, including constructing higher-quality instruction-tuning data, designing architectures tailored to streaming inputs, and incorporating stronger video encoders. More recent studies~\citep{lu2026aura, shen2026simple} show that strong VLMs with a small local video window yield surprisingly strong performance on real-time questions on existing streaming video benchmarks.

Another line of work focuses on the memory mechanisms. Some methods compress historical visual observations into textual summaries or memories~\citep{liu2026thinking, guan2026video, yao2026joyai}, while others reduce the accumulated visual tokens through clustering or token merging~\citep{zhang2024flash}. Additional approaches organize historical information into hierarchical memory structures~\citep{zeng2026streamforest, zhang2026hermes, xie2026fluxmem, liang2026oasis}, or maintain a fixed-size memory buffer that selectively preserves useful historical observations~\citep{qian2025dispider, yao2025timechat}. 
Despite their differences, these methods generally address the growing streaming context by compressing historical information into a bounded token budget. In contrast, we represent long-term memory directly in model weights.

Concurrent work~\citep{chen2026streamttt} similarly moves long-term memory outside the attention context. However, it stores historical information in a separate test-time-training (TTT) branch with online-updated fast weights. In contrast, our method directly internalizes history into the VLM itself by dynamically generating low-rank parameter adaptations, without introducing a separate memory branch.

\medskip
\noindent\textbf{Low-Rank Adaptation and Context Internalization.}
Low-rank adaptation (LoRA)~\citep{hu2021lora} has been widely adopted as a parameter-efficient approach for adapting pretrained vision-language models (VLMs) and multi-modal large language models (MLLMs) to downstream tasks or new domains~\citep{dettmers2023qlora, chen2024longlora, liu2025pave}. 
Recently, several works have explored using LoRA as a representation for encoding data-specific patterns~\citep{shrestha2025finetuning, yin2026qwen} or contextual information~\citep{charakorn2026doc, charakorn2025text}. These approaches demonstrate that model parameters can be a good representation for storing information beyond conventional token-based context. However, they typically require additional optimization when adapting to a new pattern, making it difficult to apply them directly to evolving inputs. $\delta$-mem~\citep{lei2026delta} explores dynamically representing textual information through LoRA updates without test-time training. Nevertheless, it still relies on access to the historical context during inference.

Different from these prior approaches, our work specifically handles the streaming setting, and introduces a context internalization mechanism that progressively incorporates incoming visual context into low-rank model representations. 

\section{Internalizing Long-Term History for Streaming Videos}

\subsection{Problem Formulation: Streaming Video and Internalizing History}
\label{section:method_Preliminaries}

A streaming VLM $M_\theta$, parameterized by $\theta$, must continuously perceive incoming visual content, respond to user queries or instructions $I_t$, and generate an appropriate response or decision $R_t$ at time step $t$, using information available before $t$. Formally, we define this streaming setting as
\begin{equation}
R_t = M_{\theta}(I_t, C_{0:t-1}, C_t),
\end{equation}
where $C_{0:t-1}$ denotes the historical context accumulated before time $t$, and $C_t$ denotes the current observation. Each $C$ may contain both visual and textual information. This formulation encompasses streaming video QA, VLA, and other real-time interactive settings.

Explicitly retaining $C_{0:t-1}$, however, leads to an ever-growing input context. We instead propose to \textit{internalize} this history into lightweight, evolving model parameters using LoRA weights $\Delta\theta_t$:
\begin{equation}
\Delta\theta_{t} = \Phi(C_{0:t-1}; h_0),
\qquad
R_t = M_{\theta+\Delta\theta_t}(I_t, C_t),
\end{equation}
where $\Phi$ is a recurrent HyperNetwork initialized with a fixed hidden state $h_0$, and it incrementally encodes historical context into LoRA updates. With the history-dependent $\Delta\theta_t$ applied to $M_\theta$, the VLM only needs to explicitly process the current observation $C_t$ and user query $I_t$ at inference time. Thus, \textit{history is represented as weights, while the current observation remains as input}.

\begin{figure*}[t]
    \centering
    \includegraphics[width=0.98\linewidth]{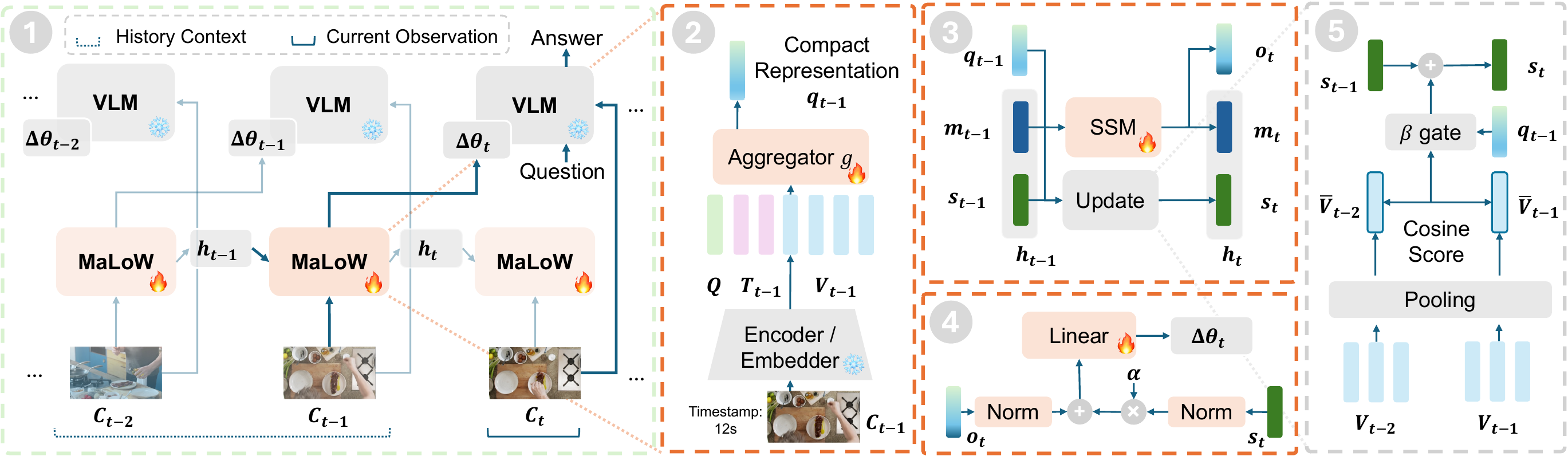}
    \vspace{-0.5em}
    \caption{\textbf{Overview of \method.} \textbf{(1)} \method internalizes history into LoRA weights applied to a base VLM that processes only the current observation. \textbf{(2)} Context encoding compresses visual tokens $V_{t-1}$ and timestamp tokens $T_{t-1}$ into a compact representation $q_{t-1}$ using a learnable query $Q$. \textbf{(3)} Hidden state updates $m_t$ that captures local dynamics and $s_t$ that preserves global semantics. \textbf{(4)} LoRA weights are decoded from the hidden states. \textbf{(5)} $s_t$ is updated through gated summation.
}
    \label{fig:Method}
    \vspace{-1em}
\end{figure*}

Importantly, we define $\Phi$ as the recurrent application of a single-step update $f$ over the history context $C_{0:t-1}$, as shown in Fig. \ref{fig:Method}-1. $f$ is itself also a HyperNetwork that updates a hidden state $h_t$ and outputs the LoRA weights $\Delta\theta_{t}$, given by
\begin{equation}
h_{t}, \ \Delta\theta_{t} = f(h_{t-1}, C_{t-1}).
\end{equation}
$h_t$ offers a compact representation of $C_{0:t-1}$ and repeated application of $f$ realizes $\Phi$.

\subsection{Our Design: \method} \label{section:our_method}
We now describe the design of \method, focusing on the single-step update $f$. Conceptually, $f$ consists of three parts: (1) the encoding of context $C_{t-1}$; (2) the update of the hidden state $h_t$; and (3) the decoding of the LoRA weights $\Delta\theta_{t}$. This is illustrated in Fig. \ref{fig:Method}-(2-5).

\medskip
\textbf{Context Encoding.}
Videos contain substantial redundancy~\citep{shang2025llava, zhong2025aim}. Given the observation $C_{t-1}$ consisting of a video clip and its textual context (e.g., timestamps in text format), we first compress it into a compact representation. This is done by first encoding the video and text into their embeddings $V_{t-1}$ (video) and $T_{t-1}$ (text) using the vision encoder and the text tokenizer of the VLM. We then learn a set of queries $Q$ to compress $V_{t-1}$ and $T_{t-1}$
\begin{equation}
q_{t-1} = g([V_{t-1}; T_{t-1}; Q])
\end{equation}
where $q_{t-1}$ is the resulting compact representation of $C_{t-1}$. $g$ is realized using a few self-attention layers, with the concatenated $[V_{t-1}, T_{t-1}, Q]$ as the input (see Fig. \ref{fig:Method}-2).

\medskip
\textbf{Hidden-State Update.}
In interactive settings, new observations arrive continuously over time. An effective memory mechanism should therefore capture both the temporal dependencies among observations and the global semantics accumulated throughout the interaction. To this end, we decompose the hidden state $h_t$ into two components: $h_t = [m_t; s_t],$
where $m_t$ models local temporal dynamics in the stream, while $s_t$ captures longer-term semantic information. 

Specifically, $m_t$ is modeled by a State-Space Model (SSM) using DeltaNet from~\citep{yang2024parallelizing}. Given $q_{t-1}$ and the previous state \(m_{t-1}\), the SSM is updated as
\begin{equation}
m_{t}, o_{t} = \mathrm{SSM}(q_{t-1}, m_{t-1}),
\end{equation}
where $m_{t}$ denotes the updated state and $o_{t}$ the corresponding output. Through recurrent updates, $m_t$ captures the temporal dependencies across the observed stream (see Fig. \ref{fig:Method}-3).

In parallel, $s_t$ is simply a gated summation of $q_{t-1}$, acting as a skip connection to allow longer-term memory~\citep{hochreiter1997long}. This is given by
\begin{equation}
s_t = s_{t-1} + \beta_{t-1} q_{t-1} \quad \text{with}\ \ \beta_{t-1} = \mathbb{I}\!\left[\text{sim}(\bar{V}_{t-1}, \bar{V}_{t-2}) \leq \beta\right], 
\end{equation}
where $\beta_t$ is a binary gate, $\mathbb{I}$ is the indicator function, and $\text{sim}(\bar{V}_{t-1}, \bar{V}_{t-2})$ is the cosine similarity on the averaged pooled $V_{t-1}$, $V_{t-2}$. This is shown in Fig. \ref{fig:Method}-5.

\textbf{LoRA Decoding.} 
Finally, we convert the accumulated historical information, represented by $o_t$ and $s_t$, into LoRA weights $\Delta\theta_{t}$ using a readout function $h$
\begin{equation}
   \Delta\theta_{t}  = h(o_t, s_t) 
\end{equation}
$h$ is realized as a linear mapping on top of $o_t$ and $s_t$ normalized with RMSNorm~\citep{zhang2019root}, with an additional learnable gating parameter $\alpha$ applied to the normalized version of $s_t$ (see Fig. \ref{fig:Method}-4). The resulting LoRA weights $\Delta\theta_{t}$ will be applied to the VLM at the next time step. 

\subsection{Model Training and Theoretic Analysis} 
\noindent\textbf{Loss Function.} We optimize \method using the loss function
\begin{equation}
  \mathcal{L}
  =
  \mathcal{L}_{\mathrm{ntp}}
  + \gamma\,\mathcal{L}_{\mathrm{LoRA}}.
\end{equation}
$\mathcal{L}_{ntp}$ is the standard next-token prediction objective
\begin{equation}
\mathcal{L}_{ntp}
=
-\mathbb{E}_{(C_{0:t-1},\, C_t,\, I_t,\, y_t)\sim\mathcal{D}}
\sum\nolimits_{k=1}^{|y_t|}
\log
p_{\theta+\Delta\theta_t}
\left(
y_{t,k}
\,\middle|\,
I_t, C_t, y_{t,<k}
\right).
\end{equation}
$\mathcal{L}_{\mathrm{LoRA}}$ regularizes the magnitude of the generated LoRA weights following the previous work~\citep{charakorn2025text}:
\begin{equation}
    \mathcal{L}_{\mathrm{LoRA}} = \Sigma_{s\in\mathcal{S}} \left( \operatorname{mean}|\mathbf{A}_s|
    +
    \operatorname{mean}|\mathbf{B}_s| \right) / |\mathcal{S}|,
\end{equation}
where $\mathcal{S}$ indexes LoRA weights attached to different parameters of the VLM, $\mathbf A_s$ and $\mathbf B_s$ are the factorized low-rank matrices in the corresponding LoRA weights. 

\medskip
\textbf{Training Scheme.} During training, VLM parameters $\theta$ is frozen and the historical context $C_{0:t-1}$ are not directly provided to the VLM when predicting $y_t$. Instead, $C_{0:t-1}$ can only influence the prediction through the generated LoRA weights $\Delta\theta_t$. The prediction loss is backpropagated through $\Delta\theta_t$ to optimize the parameters in the HyperNetwork $f$, encouraging \method to preserve historical information that is useful for future interactions while discarding redundant context.

\medskip
\textbf{Theoretical Analysis.} We also provide theoretical analysis in Appendix~\ref{appendix:math-interpretation} to establish the feasibility of representing history-induced feature transformations through low-rank parameter updates.

\subsection{Training data construction: \method-Instruct} \label{section:Training_data}
\begin{figure*}[t]
    \centering
    \includegraphics[width=0.98\linewidth]{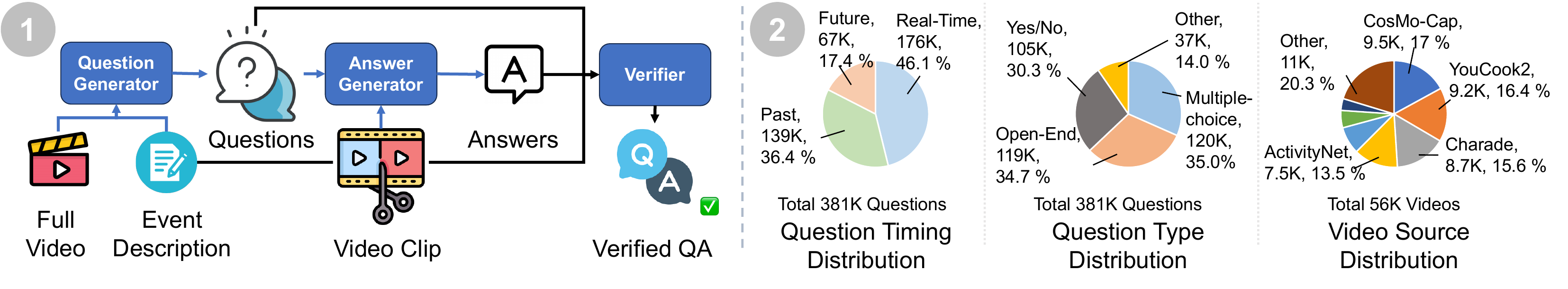}
    \vspace{-1em}
    \caption{Overview of the \method data pipeline and statistics of \textbf{\method-Instruct}. \textbf{(1)}: Our data pipeline consists of a Question Generator, an Answer Generator, and a Verifier. \textbf{(2)}: The distribution of the streaming question timing, the question type, and video source in \method-Instruct.}
    \label{fig:data_pipeline}
    \vspace{-1em}
\end{figure*}

To train \method to enable adapted VLMs to retrospect over past observations, preserve relevant information for understanding the present, and support future responses or actions, we construct a comprehensive training dataset for streaming video understanding, named as \textbf{\method-Instruct}. 

\medskip
\textbf{Data Sources and Dataset Construction.}
Our data construction pipeline consists of three main components: (1) a \emph{question generator}, (2) an \emph{answer generator}, and (3) an \emph{verifier} (Fig.\ \ref{fig:data_pipeline}-1).
We build on existing video temporal grounding and procedural action datasets that provide event/action timestamps and corresponding descriptions. For each annotated event, the question generator takes the video together with the event description and generates candidate questions about either the target event itself or its surrounding context, an \textit{expected answer}, along with the question timestamps \(t\). We consider diverse question formats, including open-ended, multiple-choice, and yes/no questions. To prevent data leakage, we filter out the samples used in OVO-Bench~\citep{niu2025ovo}.

Given a generated question, the answer generator then processes the video up to timestamp \(t\) and produces a \textit{model answer}. Finally, the verifier takes the generated question, \textit{expected answer}, \textit{model answer}, and the video context surrounding the target event to assess both the validity of the question and the correctness of the answer. An unreasonable question or an inconsistent answer typically results in disagreement between the expected answer, generated answer, and supporting video context. We therefore retain only QA pairs that pass the verification stage to form the final training dataset.
We use Qwen3.5-122B-A10B~\citep{qwen3.5} as both the question and answer generator, and Qwen3.5-397B-A17B as the verifier. Please see more details in the Appendix~\ref{appendix:data:video_source} and~\ref{appendix:data:data_pipeline}.

\medskip
\textbf{Dataset Statistics.} \method-Instruct contains 381K instruction samples on 56K videos. We summarize the distributions and statistics of the videos and questions in Fig.\ \ref{fig:data_pipeline}-2. Approximately 36\% of the QA pairs have query timestamps that occur after the corresponding event has ended, providing explicit supervision for learning the \method module to retain and retrieve historical information. Moreover, for many real-time questions, part of the relevant event falls outside the current observation window, further encouraging the model to leverage historical context through \method. The question types are approximately evenly distributed among multiple-choice, open-ended, and yes/no questions. The videos are collected from 14 datasets, among which CosMo-Cap~\citep{wang2024cosmo}, YouCook2~\citep{ZhXuCoAAAI18}, Charades-STA~\citep{gao2017tall}, and ActivityNet-Captions~\citep{krishna2017dense} together account for 62.5\% of all videos. 
More dataset statistics are presented in the Appendix~\ref{appendix:data:statistic}.

\section{Experiments and Results}\label{sec:exp}

\begin{table*}[t!]
\centering
\renewcommand{\arraystretch}{1.1}
\setlength{\tabcolsep}{3pt}

\caption{\textbf{Main results in streaming video understanding.} *VideoChat3 uses 2 fps on OVO-Bench and 4 fps on StreamingBench. \textcolor{gray}{Gray} numbers denote results reported in the original paper that differ from our re-evaluation using the official implementation and reported evaluation hyperparameters. Please refer to \textit{Baselines Re-evaluation} in sec~\ref{appendix:additional_details:main} for more information. \method (MOSS-VL-Real-11B) and \method(Qwen3-VL-4B) achieve a new state-of-the-art score of 74.5 and 84.8 on OVO-Bench and StreamingBench, respectively, while \method (Qwen3.5-9B) attains the highest EPM score of 63.6 among all compared methods. }
\vspace{-1em}
\label{table:main_result}
\medskip
\resizebox{\textwidth}{!}{
\begin{tabular}{lc|c|cccccc|c|>{\columncolor{lightgray}}ccc|c|c}
\toprule
\multirow{3}{*}{Model} & \multirow{3}{*}{\#Frames}
& \multirow{3}{*}{\shortstack{Streaming\\ Bench (R)}}
& \multicolumn{12}{c}{\textit{OVO-Bench}} \\
\cmidrule(lr){4-15}
& &
& \multicolumn{7}{c|}{\textit{Real-Time Visual Perception}}
& \multicolumn{4}{c|}{\textit{Backward Tracing}}
& \multirow{2}{*}{Avg.} \\
\cmidrule(lr){4-10} \cmidrule(lr){11-14}
& & & OCR & ACR & ATR & STU & FPD & OJR & Avg.
& \textbf{EPM} & ASI & HLD & Avg. & \\
\midrule
\multicolumn{10}{l}{\itshape\small\textbf{Offline VLMs}} \\

LLaVA-Video-7B~\citep{zhang2024llava}
& 64 & --
& 69.1 & 58.7 & 68.8 & 49.4 & 74.3 & 59.8 & 63.5
& 56.2 & 57.4 &  7.5 & 40.4
& 52.0 \\
LongVU-7B~\citep{pmlr-v267-shen25j}
& 1\,fps & --
& 55.7 & 49.5 & 59.5 & 48.3 & 68.3 & 63.0 & 57.4
& 43.1 & 66.2 &  9.1 & 39.5
& 48.5 \\
\rowcolor{lightgreen}
Qwen3-VL-4B~\citep{bai2025qwen3} & 
1 fps 
& 78.4
& 85.2 & 60.6 & 74.1 & 51.1 & 76.2 & 67.9 & 69.2 
& 57.9 & 61.5 & 16.7 & 45.4 & 57.3
\\
\rowcolor{lightgreen}
Qwen3.5-9B~\citep{qwen3.5} & 
1 fps 
& 81.9 
& 83.9 & 62.4 & 70.7 & 46.1 & 76.2 & 70.6 & 68.3 
& 61.6 & 62.8 & 59.1 & 61.2 & 64.8 \\
\midrule

\multicolumn{10}{l}{\itshape\small\textbf{Streaming VLMs with Long-term (Full) Context}} \\
VideoLLM-online-8B~\citep{chen2024videollm}
& 2\,fps & 36.0
&  8.1 & 23.9 & 12.1 & 14.0 & 45.5 & 21.2 & 20.8
& 22.2 & 18.8 & 12.2 & 17.7
& 19.3 \\
Streamo-7B~\citep{xia2026streaming}
& 1 fps & --
& 79.2 & 57.8 & 75.0 & 49.4 & 64.4 & 70.1 & 66.0
& 54.6 & 52.0 & 31.7 & 46.1
& 56.1 \\
\rowcolor{lightblue}
VideoChat3-4B~\citep{li2026videochat3} 
& 2-4 fps* & 79.0 \textcolor{gray}{(83.0)}
& 92.6 & 78.0 & 77.6 & 59.0 & 74.3 & 71.7 & 75.5
& 58.8 & 58.6 & 46.8 & 54.7
& 65.1 \\
\rowcolor{lightyellow}
MOSS-VL-Real-11B~\citep{wang2026moss} 
& 1 fps & 80.6 \textcolor{gray}{(82.9)}
& 86.6 & 70.6 & 84.8 & 64.6 & 71.2 & 73.9 & 75.2
& 60.6 & 64.5 & 92.4 & \textbf{72.5} 
& 73.8 \\

\multicolumn{10}{l}{\itshape\small\textbf{Streaming VLMs with Local Context Window}} \\
Mage-VL-4B~\citep{yang2026mage} 
& 4 (1 fps) & -- 
& 94.6 &  83.5 &  79.3 &  74.2 &  70.3 &  77.2 
&  79.8  & 50.2 &  61.5 &  32.8 &  48.2 
&  64.0 \\
Simplestream-8B~\citep{shen2026simple} 
& 4 (1 fps) & 78.3
& 89.3 & 77.1 & 83.6 & 68.5 & 76.2 & 81.0 & 79.3
& 49.5 & 56.1 & 54.8 & 53.5
& 66.4 \\
AURA-8B~\citep{lu2026aura} 
& 30 (1 fps) & 83.2
& 89.9 & 79.8 & 80.2 & 70.2 & 77.2 & 81.5 & \textbf{79.8} 
& 54.9 & 67.6 & 58.6 & 60.4 
& 70.1 \\

\multicolumn{10}{l}{\itshape\small\textbf{Streaming VLMs with Local Context Window and Memory Mechanism}} \\
Flash-VStream-7B~\citep{zhang2024flash}
& 1\,fps & 23.2
& 24.2 & 29.4 & 28.5 & 33.7 & 25.7 & 28.8 & 28.4
& 39.1 & 37.2 &  5.9 & 27.4
& 27.9 \\
Dispider-7B~\citep{qian2025dispider}
& 1\,fps & 67.6
& 57.7 & 49.5 & 62.1 & 44.9 & 61.4 & 51.6 & 54.6
& 48.5 & 55.4 &  4.3 & 36.1
& 45.4 \\
TimeChat-Online-7B~\citep{yao2025timechat}
& 1\,fps & 75.3
& 75.2 & 46.8 & 70.7 & 47.8 & 69.3 & 61.4 & 61.9
& 55.9 & 59.5 &  9.7 & 41.7
& 51.8 \\
StreamForest-7B~\citep{zeng2026streamforest}
& 1 fps & 77.3
& 68.5 & 53.2 & 71.6 & 47.8 & 65.4 & 60.9 & 61.2
& 58.9 & 64.9 & 32.3 & 52.0
& 56.6 \\
JoyAI-VL-Interaction-8B~\citep{yao2026joyai} 
& 1 fps & -- 
& 72.5 & 61.5 & 75.0 & 59.0 & 78.2 & 64.1 & 68.4 
& 62.0 & 70.9 & 12.9 & 48.6 
& 58.5 \\
HERMES-7B~\citep{zhang2026hermes}
& 1 fps & 79.4
& 85.2 & 64.2 & 71.6 & 53.4 & 74.3 & 65.2 & 69.0
& 48.5 & 62.2 & 37.6 & 49.4
& 59.2 \\
ThinkStream-3B~\citep{liu2026thinking} 
& 2 fps & 75.0 
& 85.2 &  64.2 & 69.8& 49.4& 69.3& 64.1& 67.0
& 53.9& 59.5& 43.6& 52.3
& 59.7 \\
VST-7B~\citep{guan2026video} 
& 2 fps & 79.5 
& 80.5 & 55.1 & 72.4 & 55.1 & 76.2 & 64.1 & 67.2 
& 56.9 & 64.9 & 48.4 & 56.7 
& 62.0 \\
OASIS-8B~\citep{liang2026oasis} 
& 2 fps & 78.2
& 92.0 & 80.7 & 81.0 & 67.4 & 67.3 & 79.9 & 78.1 
& 62.0 & 60.1 & 47.3 & 57.2 & 67.7 \\
StreamTTT-4B~\citep{chen2026streamttt} 
& 2 fps & 81.3 
& 91.3 & 80.7 & 80.2 & 56.2 & 82.2 & 77.7 & 78.1 
& 60.9 & 58.1 & 60.8 & 59.9 
& 69.0 \\
\midrule
\multicolumn{10}{l}{\itshape\small\textbf{Ours}} \\
\rowcolor{lightblue}
\method (VideoChat3-4B) - 197.6M trainable (4.0\%)
& 1 fps 
& 81.2
& 89.9 & 70.6 & 80.2 & 60.7 & 76.2 & 75.3 & 75.5 
& 60.2 & 58.1 & 54.8 & 57.7 & 66.6 \\

\rowcolor{lightgreen}
\method (Qwen3-VL-4B) - 197.6M trainable (3.9\%)
& 1 fps 
& \textbf{84.8}
& 90.6 & 69.7 & 75.0 & 64.0 & 74.3 & 74.5 & 74.7
& 62.0 & 67.6 & 61.8 & 63.8 & 69.3 \\ 

\rowcolor{lightgreen}
\method (Qwen3.5-9B) - 308.9M trainable (3.0\%)
& 1 fps 
& 82.3
& 87.9 & 66.1 & 81.0 & 60.1 & 84.2 & 77.7 & 76.2 
& \textbf{63.6} & 66.2 & 69.4 & 66.4 & 71.3 \\ 

\rowcolor{lightyellow}
\method (MOSS-VL-Real-11B) - 321.1M trainable (1.4\%)
& 1 fps 
& 81.0
& 91.9 & 74.3 & 85.3 & 62.9 & 73.3 & 76.6 & 77.4 
& 62.0 & 62.8 & 90.3 & 71.7 & \textbf{74.5} \\ 

\bottomrule
\end{tabular}
}
\vspace{-2mm}
\end{table*}

\subsection{Streaming Video Understanding} \label{sec:main_exp}
\textbf{Benchmarks.}
Following the prior works~\citep{shen2026simple, zhang2026hermes},
we evaluate \method on the \textit{Real-Time} split of StreamingBench~\citep{lin2026streamingbench}, as well as the \textit{Real-Time} and \textit{Backward} splits of OVO-Bench~\citep{niu2025ovo}. 
The \textit{Backward} splits of OVO-Bench mostly focus on retrospect to the past and comprise three tasks: Episodic Memory (EPM), Action Sequence Identification (ASI), and Hallucination Detection (HLD). Among them, EPM is particularly well suited for assessing the effectiveness of the memory mechanism introduced by \method.

\medskip
\textbf{Experiment Protocol.} We build \method on top of a diverse set of VLMs, including models pre-trained for offline video understanding (Qwen3-VL-4B~\citep{bai2025qwen3} and Qwen3.5-9B~\citep{qwen3.5}) and models designed for streaming video (VideoChat3-4B~\citep{li2026videochat3} and MOSS-VL-Realtime-11B~\citep{wang2026moss}). Across all base VLMs, \method operates in a \textit{streaming setting} with a local context window: at each time step $t$, only the most recent video clip $C_t$ is provided as the current observation, while all preceding clips $C_{0:t-1}$ are consolidated into the evolving model weights.


\medskip
\textbf{Implementation Details.} We apply rank-8 LoRA to the MLP in each attention layer and \textit{v\_proj} layers in each self-attention layer. For Qwen3.5-9B, we additionally apply LoRA to the \textit{in\_proj\_qkv\_proj} layers in its linear-attention blocks. For MOSS-VL, we additional apply \textit{v\_proj} layers in each cross-attention layer. 
We use a four-layer self-attention network as the aggregator, with a hidden dimension of 512 and eight attention heads. For each layer of the base model, we learn an independent query and maintain a corresponding hidden state. We instantiate the SSM using a four-layer DeltaNet~\citep{yang2024parallelizing} with a hidden dimension of 512. To generate the LoRA parameters, we learn separate linear projection layers for each target module and LoRA rank component. 
We train \method on the training data introduced in Sec.\ \ref{section:Training_data} for 2,800 steps with a total batch size of 64. We use the Adam optimizer with a peak learning rate of $8\times10^{-5}$ and a cosine learning-rate schedule.
The videos are sampled at 1 fps during training and inference of \method. Each $C_t$ is a 2-second video clip. 
Additional implementation details are provided in the Appendix~\ref{appendix:additional_details:main}.

\medskip
\textbf{Baselines.} We categorize the baselines into four groups according to the model types and how they manage historical visual context during streaming inference.
\begin{enumerate}

\item \textbf{Offline VLMs.} This category includes LLaVA-Video-7B~\citep{zhang2024llava}, LongVU-7B~\citep{pmlr-v267-shen25j}, Qwen3-VL-4B~\citep{bai2025qwen3}, and Qwen3.5-9B~\citep{qwen3.5}. We compare to these models for their strong video understanding capabilities, acquired through large-scale video-text pretraining. We evaluate their pretrained checkpoints by treating streaming videos as offline inputs.

\item \textbf{Streaming VLMs with Long-term (Full) Context.}
This category includes VideoLLM-Online~\citep{chen2024videollm}, Streamo~\citep{xia2026streaming}, VideoChat3~\citep{li2026videochat3}, and MOSS-VL-Realtime~\citep{wang2026moss}. These methods preserve historical observations directly in the inference context as memory for streaming video understanding, and for short videos often retaining the entire video history. As the stream grows, this leads to an ever expanding context window and substantial memory and computational overhead. We note that VideoChat3 uses a local context window on OVO-Bench, while retaining the full historical context on StreamingBench.


\item \textbf{Streaming VLMs with a Local Context Window.}
This category includes Mage-VL~\citep{yang2026mage}, SimpleStream~\citep{shen2026simple}, and AURA~\citep{lu2026aura}. We compare to these models for their strong real-time performance, achieved by retaining only the most recent observations within a fixed local window of approximately 4–30 seconds. Historical observations that fall outside the local window are evicted~\citep{xu2026streamingvlm}.

\item \textbf{Streaming VLMs with a Local Context Window and Memory Mechanism.}
This category includes Flash-VStream~\citep{zhang2024flash}, Dispider~\citep{qian2025dispider}, TimeChat-Online~\citep{yao2025timechat}, StreamForest~\citep{zeng2026streamforest}, JoyAI-VL~\citep{yao2026joyai}, HERMES~\citep{zhang2026hermes}, ThinkStream~\citep{liu2026thinking}, VST~\citep{guan2026video}, OASIS~\citep{liang2026oasis} and StreamTTT~\citep{chen2026streamttt}. Similar to our design, these methods retain most recent observations in a local context window while compressing historical information into a more compact memory representation, such as textual summaries or condensed visual features. Among them, the concurrent work StreamTTT is most relevant. StreamTTT stores historical information in an online-updated test-time-training (TTT) branch while retaining the most recent visual tokens as the current observation.

\end{enumerate}

\medskip
\textbf{\method achieves state-of-the-art results.} As shown in Table~\ref{table:main_result}, \method establishes strong performance across streaming video benchmarks. \method (MOSS-VL-Realtime-11B) and \method (Qwen3-VL-4B) achieve new state-of-the-art scores of 74.5\% and 84.8\% on OVO-Bench and StreamingBench, respectively, while \method (Qwen3.5-9B) attains the highest EPM score of 63.6\% among all compared methods.
 
We highlight key comparisons at matched model scales. \textit{At the 4B scale}, the \textcolor{darkgreen}{\method (Qwen3-VL-4B)} surpasses the concurrent work StreamTTT-4B~\citep{chen2026streamttt} by 3.9\% and 3.5\% on the backward split of the OVO-bench and StreamingBench, respectively. 
\textit{At the 7-9B scale}, the \textcolor{darkgreen}{\method (Qwen3.5-9B)} outperforms the previous memory-based method OASIS-8B~\citep{liang2026oasis} and local window-based method AURA-8B~\citep{lu2026aura} by 9.2\% and 6.0\% on the backward split of the OVO-bench, respectively.
\textit{At the 11B scale}, \textcolor{darkyellow}{\method (MOSS-VL-Real-11B)} improves over its base MOSS-VL-Real-11B by 2.2\% and 1.4\% on the Real-Time split and EPM task of OVO-Bench, respectively.
We recognize that on the OVO-Bench real-time split, \textcolor{darkgreen}{\method (Qwen3-VL-4B)} and \textcolor{darkgreen}{\method (Qwen3.5-9B)} underperform StreamTTT-4B and AURA-8B, respectively. We hypothesize that these baselines may have benefited from fine-tuning their base VLMs, whereas \method keeps the base VLMs intact.

Collectively, these results show that \method is particularly effective on memory-critical tasks while retaining competitive performance on real-time perception. Importantly, the gains are consistent across model scales from 4B to 11B and across base VLMs pre-trained for both offline video understanding (Qwen series) and streaming video understanding (VideoChat3 and MOSS-VL-Realtime). 


\medskip
\textbf{\method offers a lightweight memory patch.} Across different base VLMs, \method introduces only 1.4\%-4.0\% additional parameters. The pretrained backbone architecture and parameters remain unchanged, while historical information is encoded entirely through the evolving LoRA weights. Preserving the backbone helps retain its pretrained capabilities and  minimizes interference with real-time perception. Moreover, widespread LoRA support in modern inference frameworks facilitates efficient deployment.



\medskip
\textbf{\method incurs minor inference overhead.} We profile the streaming inference latency of \method (Qwen3.5-9B) and four baselines on 100 deterministically sampled questions from OVO-Bench. At each query timestamp $t$, we measure the time required to process the current observation, prefill the KV cache, and decode the first token. For \method, we additionally include the cost of generating and applying the LoRA weights. Further experimental details are provided in Appendix~\ref{appendix:additional_details}.
As shown in Fig.\ \ref{fig:speed}, \method achieves a favorable trade-off between inference latency and performance on memory-critical tasks. The additional latency introduced by our memory mechanism is only around 23 ms, accounting for less than 15\% of the total inference time (Table~\ref{tab:c2l-latency-breakdown} in the Appendix). Moreover, LoRA generation can be parallelized with backbone inference, providing further opportunities to reduce the runtime overhead.

%


\medskip
\textbf{\method excels at long-horizon memory tasks.} We further stress-test our memory mechanism on the Historical Retrospection (HR) split of StreamArena~\citep{zhang2026streamarena}; implementation details are provided in Appendix~\ref{appendix:additional_details}. HR specifically targets long-horizon memory: the query is issued substantially later than the appearance of the relevant visual evidence, with a median query-evidence gap of 12.1 minutes. We compare \method (Qwen3.5-9B) with strong streaming video models, including MOSS-VL-Realtime-11B~\citep{wang2026moss}, VideoChat3-4B~\citep{li2026videochat3}, AURA-8B~\citep{lu2026aura}, and VST-7B~\citep{guan2026video}. These models either maintain a long context window (MOSS-VL-Realtime), or compress historical observations into text-based memory (VideoChat3, AURA, and VST).
As reported in Table~\ref{tab:StreamArena_hr}, \method achieves an average score of 22.8, outperforming VST and MOSS-VL-Realtime by 2.2\% and 4.5\%, respectively. These results highlight the effectiveness of representing history in evolving model weights for retaining information over long temporal horizons.


\textbf{\method generalizes across model scales and families.}
We further evaluate \method across a broader range of model scales (0.8B--4B) and model families (Qwen3.5 and Gemma4); implementation details are provided in Appendix~\ref{appendix:additional_details}. Following the official benchmark protocols, we evaluate all three splits of OVO-Bench (Backward, Real-Time, and Forward) and all four splits of StreamingBench (Real-Time, Omni, Proactive, and SQA). For each backbone, we compare \method against two baselines: (1) the corresponding base model operating with full context and (2) the same model restricted to a local context window.
Table~\ref{tab:different_scale_diffent_source} shows that \method consistently improves overall performance \textit{across model scales and families}. In particular, \method outperforms the full-context baseline on real-time perception splits (\textcolor{green!50!black}{green cells}), demonstrating that \method effectively alleviates the negative effect of history on real-time perception. Meanwhile, \method substantially improves over the local-window baseline on memory-intensive splits (\textcolor{blue!60!black}{blue cells}), highlighting the effectiveness of the internalized memory. 



\subsection{Additional Demonstration and Analysis}

\begin{figure*}[t]
    \centering

    \begin{minipage}[t]{0.49\textwidth}
        \vspace{0pt}
        \centering
        \vspace{-0.5em}
        \captionof{table}{
            \textbf{VLA results} on SimplerEnv-Bridge~\citep{li24simpler}
            with the WidowX robot.
            SGB: Stack Green Block, PC: Put Carrot,
            PS: Put Spoon, PE: Put Eggplant.
        }
        \scalebox{0.72}{
        \begin{tabular}{lccccc}
            \toprule
            Method & SGB & PC & PS & PE & AVG \\
            \midrule
            Q3.5-2B-MLP
                & 20.8 & 75.0 & 70.8 & 70.8 & 59.4 \\
            \method (Q3.5-2B-MLP)
                & \textbf{29.2} & \textbf{75.0} & \textbf{87.5}
                & \textbf{83.3} & \textbf{68.8} \\
            \bottomrule
        \end{tabular}
        }
        \label{tab:vla}
        
        \vspace{-0.1em}
        \captionof{table}{
            \textbf{Results in long-horizon memory tasks} on the HR
            split of StreamArena.
        }
        \scalebox{0.72}{
        \begin{tabular}{lccccc}
            \toprule
            Method & Avg. & L1 & L2 & L3 & L4 \\
            \midrule
            VideoChat3-4B
                & 18.0 & 16.6 & 21.8 & 17.9 & 12.6 \\
            MOSS-VL-Realtime-11B
                & 18.3 & 17.1 & 22.7 & 16.3 & 13.7 \\
            VST-7B
                & 20.6 & \textbf{24.3} & 21.8
                & \textbf{19.6} & 13.2 \\
            AURA-8B
                & 21.4 & 21.8 & 26.4 & 19.3 & 14.0 \\
            \midrule
            \method (Qwen3.5-9B)
                & \textbf{22.8} & 24.0 & \textbf{28.1}
                & 19.2 & \textbf{15.7} \\
            \bottomrule
        \end{tabular}
        }
        
        \label{tab:StreamArena_hr}

    \end{minipage}
    \hfill
    \begin{minipage}[t]{0.49\textwidth}
        \vspace{0pt}
        \centering
        \vspace{-0.5em}
        \includegraphics[width=0.85\linewidth]{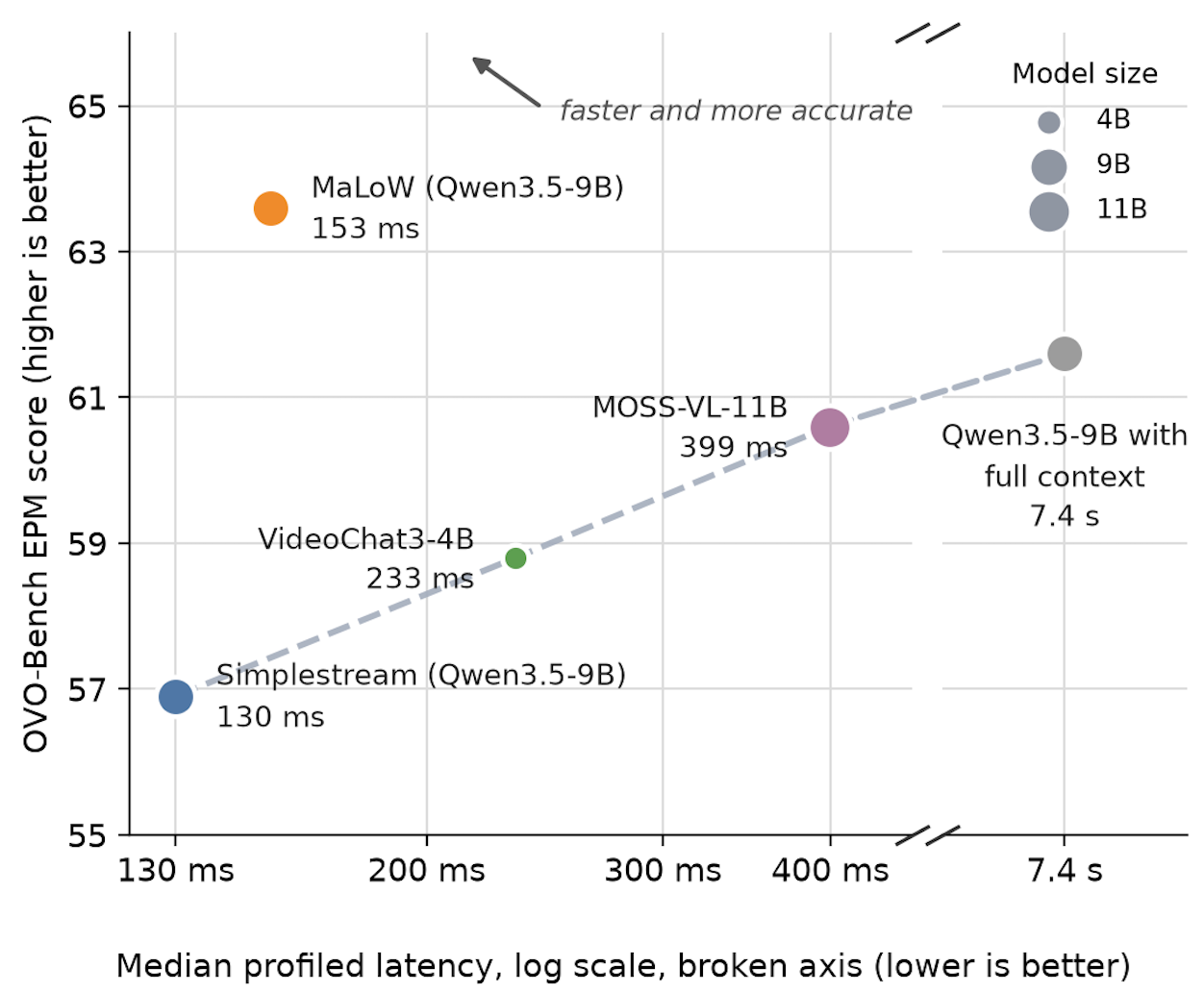}
        \vspace{-1em}
        \caption{
            \textbf{Latency--accuracy tradeoffs} of \method and baselines on the Episodic
            Memory (EPM) task of OVO-Bench. 
        }
        \label{fig:speed}

    \end{minipage}

    \vspace{-4.5mm}
\end{figure*}

\medskip
\textbf{Extension to Vision-Language-Action (VLA) Models.}
We further extend \method to robotic planning, where it is used to internalize historical visual observations and action trajectory for action prediction. Following VLM4VLA~\citep{zhang2026vlmvla}, we adapt Qwen3.5-2B~\citep{qwen3.5} by attaching a lightweight MLP action decoder and fine-tuning the VLM backbone, resulting in our baseline model, \textbf{Q3.5-2B-MLP}. The model is trained on the Bridge and RT-1 subsets of the OXE dataset~\citep{o2024open} to predict future actions based solely on the current visual observation.
We then train \method on top of Q3.5-2B-MLP using the Bridge subset of OXE. During this stage, the VLM backbone is frozen and receives only the current observation, while \method and the MLP action decoder are optimized to internalize historical observations and action trajectories and produce more accurate action trajectories. We refer to the resulting model as \textbf{\method (Q3.5-2B-MLP)}. 
We evaluate both models on the Simpler-Bridge with WidowX robot, with results reported in Table~\ref{tab:vla}. \method improves the average success rate from 59.4\% to 68.8\%, corresponding to a major gain of \textbf{9.4\%}. These results demonstrate that \method can effectively internalize historical information and leverage it to refine action trajectories without explicitly providing them to the base model. Appendix~\ref{appendix:additional_details:vla} provides more implementation details.

\begin{figure*}[t]
    \centering
    \vspace{-0.3em}
    \includegraphics[width=0.95\linewidth]{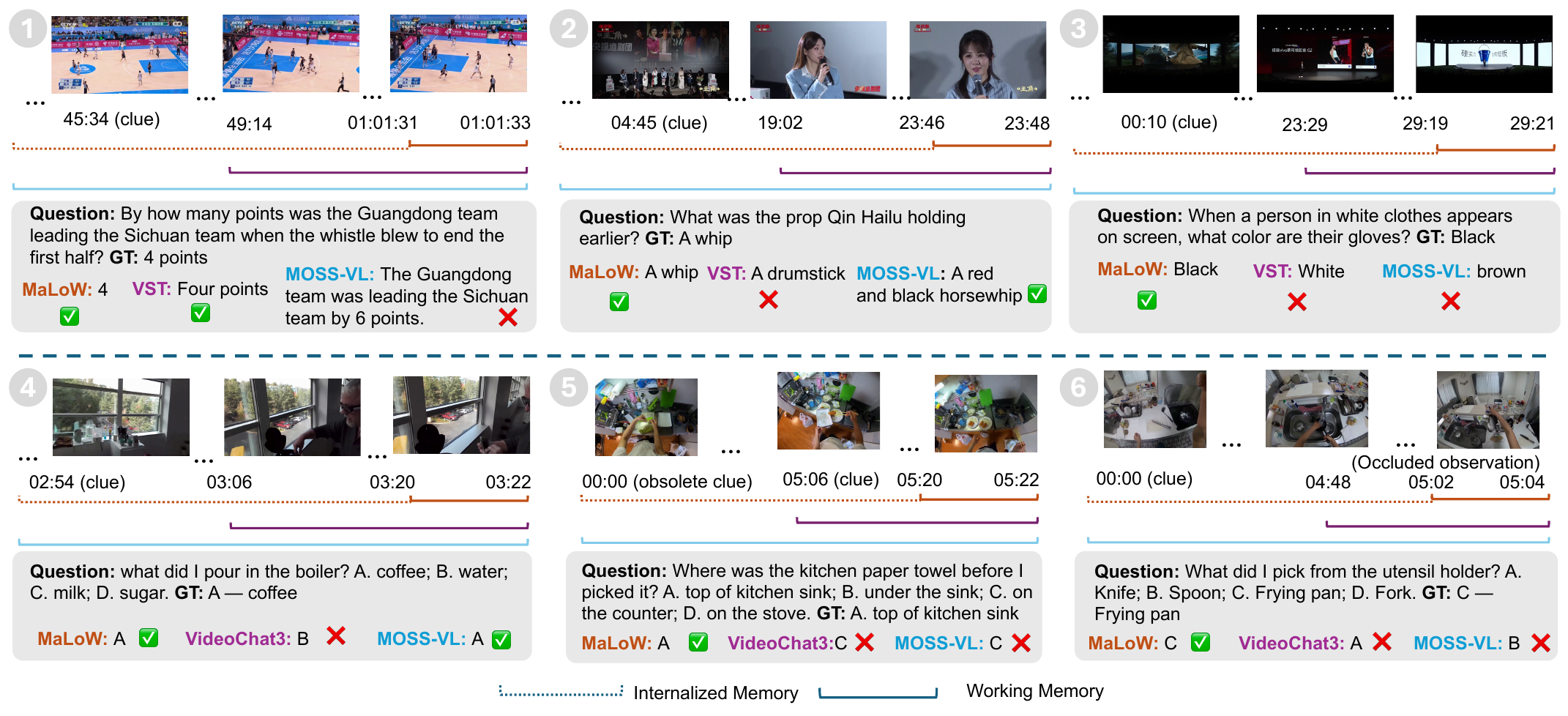}
    \vspace{-0.8em}
    \caption{\textbf{Visualization of results.}  Top row (1-3): results from StreamArena (HR) split. Bottom row (4-6): results from OVO-Bench (EPM) task. We show results from \method (Qwen3.5-9B), VideoChat3-4B,  and MOSS-VL-11B. 
    }
    
    \label{fig:visualization}
    \vspace{-4mm}
\end{figure*}

\medskip
\textbf{Qualitative Results.} We present qualitative results in Fig.\ \ref{fig:visualization}, with examples from StreamArena (HR) and OVO-Bench (EPM) shown in the top and bottom rows, respectively. We compare the results from \method (Qwen3.5-9B) with VST-7B, VideoChat3-4B, and MOSS-VL-realtime-11B. Overall, \method produces more accurate answers than the baselines on memory-intensive tasks, demonstrating the effectiveness of its history internalization mechanism. Notably, samples 1, 2, and 3 demonstrate that \method can recall the details from 16 minutes, 19 minutes, and 29 minutes ago, respectively. Sample 5 shows that \method can continuously update its memory and avoid being misled by obsolete cues. Sample 6 further demonstrates that the internalized memory complements the current observation, enabling the model to recover relevant information even when the current visual evidence is occluded. See our project website for video demonstrations. 

\begin{table*}[t]
\centering
\vspace{-0.5em}
\caption{
 \textbf{Results across base model scales and families}. Green: comparison to full-context baselines on real-time perception tasks; Blue: comparison to local-window baselines on memory-intensive tasks. 
}
\scalebox{0.72}{
\begin{tabular}{lcccc|ccccc}
\toprule
\multirow{2}{*}{Method}
& \multicolumn{4}{c|}{OVO-Bench~\citep{niu2025ovo}}
& \multicolumn{5}{c}{StreamingBench~\citep{lin2026streamingbench}} \\
& Back & Real & Forw & Avg.
& Real & Omni & Pro & SQA & Avg. \\
\midrule

\multicolumn{10}{l}{\itshape\small\textbf{Base model with 1 fps, full-context baseline}} \\
Gemma4-E4B~\citep{team2026gemma}
& 56.8 & \greenhl{63.1} & 37.9 & 52.6
& \greenhl{68.6} & 37.0 & 24.4 & 50.0 & 54.6 \\

Qwen3.5-0.8B
& 34.2 & \greenhl{60.8} & 48.0 & 47.7
& \greenhl{69.1} & 32.1 & 22.4 & 39.6 & 52.5 \\

Qwen3.5-2B
& 44.6 & \greenhl{62.5} & 46.3 & 51.2
& \greenhl{75.1} & 38.1 & 6.4 & 42.8 & 57.2 \\

Qwen3.5-4B
& 60.1 & \greenhl{66.9} & 51.2 & 59.4
& \greenhl{80.4} & 40.7 & 54.8 & 46.0 & 63.8 \\

\midrule
\multicolumn{10}{l}{\itshape\small\textbf{Base model with the last 2 frames from the last 2 seconds, local-window baseline}} \\

Gemma4-E4B
& \bluehl{54.9} & 65.7 & 39.6 & 53.4
& 67.1 & 51.1 & 20.4 & 46.8 & 58.1 \\

Qwen3.5-0.8B
& \bluehl{32.3} & 69.9 & 45.1 & 49.1
& 70.1 & 53.5 & 18.8 & 48.0 & 60.5 \\

Qwen3.5-2B
& \bluehl{39.2} & 72.8 & 45.1 & 52.4
& 76.7 & 54.2 & 5.2 & 56.8 & 64.1 \\

Qwen3.5-4B
& \bluehl{50.6} & 75.1 & 39.3 & 55.1
& 79.4 & 60.9 & 45.6 & 52.8 & 69.8 \\

\midrule
\multicolumn{10}{l}{\itshape\small\textbf{\method models with 1 fps, maximum 768 frames}} \\

\method (Gemma4-E4B) -- 190.4M trainable (1.2\%)
& \bluehl{62.6}
& \greenhl{64.8}
& 47.8
& \textbf{58.4}
& \greenhl{74.6}
& 44.8
& 40.0
& 46.0
& \textbf{61.1} \\

\method (Qwen3.5-0.8B) -- 115.5M trainable (10.8\%)
& \bluehl{54.7}
& \greenhl{67.7}
& 48.7
& \textbf{57.0}
& \greenhl{76.7}
& 48.9
& 26.0
& 48.4
& \textbf{63.0} \\

\method (Qwen3.5-2B) -- 170.7M trainable (6.3\%)
& \bluehl{63.1}
& \greenhl{70.1}
& 49.3
& \textbf{60.9}
& \greenhl{80.3}
& 54.3
& 54.4
& 55.6
& \textbf{68.8} \\

\method (Qwen3.5-4B) -- 235.0M trainable (4.6\%)
& \bluehl{62.5}
& \greenhl{74.7}
& 49.3
& \textbf{62.2}
& \greenhl{82.3}
& 63.7
& 68.0
& 50.0
& \textbf{73.5} \\
\bottomrule
\end{tabular}
}
\vspace{-1mm}
\label{tab:different_scale_diffent_source}
\end{table*}

\medskip
\textbf{Additional Ablation Studies and Analyses.}
We provide ablation studies in Section~\ref{appendix:ablation_study} of the Appendix. Similarly, additional analyses of latency, generated LoRA adapters, and qualitative failure cases can be found in Section~\ref{appendix:additional_analysis} of the Appendix.

\section{Conclusion}
In this work, we presented \textbf{\method}, a new memory framework for streaming VLMs. Rather than repeatedly injecting compressed or retrieved history into the input context, \method represents \textit{history as LoRA weights while keeping the current observation as input}. This separation provides a natural way to carry long-term information forward without revisiting the full video stream or continually expanding the context window. It also yields \textit{a lightweight memory patch}: the base VLM architecture and parameters remain unchanged, while only the compact LoRA weights evolve over time.
To realize this design, we developed a recurrent HyperNetwork that incrementally updates LoRA weights, together with \textbf{\method-Instruct}, a large-scale streaming video dataset containing 381K QA pairs over 56K videos. \method achieves strong results across streaming video understanding benchmarks, with substantial gains on memory-critical tasks and only modest inference overhead. We further show that the same memory mechanism transfers to VLA models, where incorporating historical context improves action success. We hope that \method offers a useful perspective on long-term memory for streaming VLMs: beyond asking what information should be stored or retrieved, we can also reconsider \textit{where memory should reside}, in this case, directly within evolving model weights.

\bibliography{iclr2027_conference}
\bibliographystyle{bibstyle}

\clearpage
\clearpage
\appendix
\section*{Appendix: Supplementary Materials}

\subsection*{Overview}
This appendix provides the mathematical interpretation, dataset documentation, implementation details, and additional empirical results supporting \method. The materials are organized as follows:
\begin{itemize}[leftmargin=*, itemsep=0.5em]
    \item \textbf{Appendix~\ref{appendix:math-interpretation}: Mathematical Interpretation.}
    Historical context as a read operator, its realization through low-rank MLP updates, online memory updates, and extensions to normalized attention and backbone computation.
    \item \textbf{Appendix~\ref{appendix:data}: Additional Dataset Details.}
    Video sources, the data pipeline and annotation construction, and statistics of \method-Instruct.
    \item \textbf{Appendix~\ref{appendix:additional_details}: Additional Experiment Details.}
    Model architecture, training and evaluation settings, the vision-language-action setting, different VLM scales and sources, the memory stress test, and streaming inference speed profiling.
    \item \textbf{Appendix~\ref{appendix:additional_experiments}: Additional Experiments.}
    Ablations of LoRA placement and rank, memory architecture, observation windows, and state-update frequency; analyses of latency, hidden-state evolution, and real-time perception; and additional qualitative results.
\end{itemize}

\providecommand{\R}{\mathbb{R}}
\providecommand{\norm}[1]{\left\lVert#1\right\rVert}
\providecommand{\rank}{\operatorname{rank}}
\providecommand{\pending}[1]{\par\smallskip\noindent{\color{red}\textbf{To validate.} #1}\par\smallskip}
\numberwithin{equation}{section}

\section{Mathematical Interpretation}
\label{appendix:math-interpretation}

\method carries information from earlier video clips through history-dependent parameter updates. This appendix provides a mathematical interpretation of the feasibility of encoding the effect of historical context into low-rank updates of the MLP projections within a Transformer block. Specifically, we trace how a query reads historical information in section~\ref{appendix:math:hist-read}, how this read induces a transformation of the features entering the MLP in section~\ref{appendix:math:feature-transform}, and how such a transformation can be realized or approximated by LoRA-style weight updates in section~\ref{appendix:math:mlp-realization}. The analysis does not claim that \method explicitly reconstructs the analytical history matrix \(M_H\) or is equivalent to full-context attention. Rather, it establishes that historical information can, under the assumptions described below, be represented in parameter space through low-rank updates. \method can be viewed as one of the learned instantiations of this principle, with our empirical results demonstrating that its recurrent state and generated adapters effectively preserve history-dependent information in practice. The exact identities below first assume fixed historical keys and values and an unnormalized dot-product read, while Section~\ref{appendix:math:online-update} and Section~\ref{appendix:math:backbone-extensions} extend the interpretation to the online setting, normalized attention, and the normalization operations in the backbone.

\subsection{Historical Context as a Read Operator}
\label{appendix:math:hist-read}

A streaming video arrives as a sequence of clips. At step $t$, let
$H=(C_1,\ldots,C_{t-1})$ denote the clips already observed. Suppose their visual
tokens have produced $n_H$ key--value pairs
$k_i\in\R^{d_k}$ and $v_i\in\R^{d_v}$. A current query $q\in\R^{d_k}$ reads this
history through
\begin{equation}
  O_H(q)=\sum_{i=1}^{n_H} v_i(k_i^\top q).
  \label{eq:app-token-read}
\end{equation}
All vectors are columns. Keys and values are fixed while the query varies.

Each term first measures a match, $k_i^\top q$, and then weights the corresponding
value $v_i$. Collecting the terms gives
\begin{equation}
  O_H(q)=M_Hq,
  \qquad
  M_H=\sum_{i=1}^{n_H}v_i k_i^\top
  \in\R^{d_v\times d_k}.
  \label{eq:app-history-matrix}
\end{equation}
Thus the same read can be computed either from historical tokens or from one
matrix. The matrix stores how the history responds to a query: $M_H$ is shared
by all queries for that fixed history. Its dimensions depend on the feature
widths, while the number of stored tokens grows with the video. This is the
associative-memory interpretation of linear attention
\citep{katharopoulos2020transformers,schlag2021linear}.

\subsection{From Historical Reads to Feature Transformations}
\label{appendix:math:feature-transform}

To connect the read to an MLP, consider a block with current feature
$x\in\R^d$, query projection $W_Q\in\R^{d_k\times d}$, and read-output projection
$W_O\in\R^{d\times d_v}$. The same feature supplies the query and the residual
stream:
\begin{equation}
  q=W_Qx,\qquad
  x_H=x+W_OO_H(q),\qquad
  y_H=F(x_H),
  \label{eq:app-analysis-block}
\end{equation}
where $F$ is the MLP. The feature entering the MLP changes from $x$ to $x_H$
when the historical read is added.

Substituting Eq.~\eqref{eq:app-history-matrix} into this block gives
\begin{equation}
  x_H=(I+P_H)x,
  \qquad
  \boxed{P_H=W_OM_HW_Q}\in\R^{d\times d}.
  \label{eq:app-feature-transform}
\end{equation}
$M_H$ maps a query to a historical read. $P_H$ expresses that same read as a
change to the MLP input. The projections $W_Q$ and $W_O$ determine their
relation. Consequently, the output to reproduce is
\begin{equation}
  y_H=F\bigl((I+P_H)x\bigr).
  \label{eq:app-target}
\end{equation}
This identifies a concrete role for an MLP update: reproduce how the MLP would
process the feature after history has changed it.

\subsection{Low-Rank Realizations in MLP Projections}
\label{appendix:math:mlp-realization}

Let a rank-$r$ factorization approximate the history matrix:
\begin{equation}
  \widehat M_H=B_HA_H,\qquad
  B_H\in\R^{d_v\times r},\quad A_H\in\R^{r\times d_k}.
  \label{eq:app-low-rank-read}
\end{equation}
The corresponding feature transformation is
\begin{equation}
  \widehat P_H=W_OB_HA_HW_Q,
  \qquad \rank(\widehat P_H)\leq r.
  \label{eq:app-low-rank-feature}
\end{equation}
These factors are shared across queries for the same history. The required rank
depends on how accurately they reproduce the reads used by the model.

\paragraph{Gate and up Projections.}
Consider the gated MLP
\begin{equation}
  F(x)=W_d\bigl[\sigma(W_gx)\odot(W_ux)\bigr],
  \label{eq:app-gated-mlp}
\end{equation}
where $W_g,W_u\in\R^{m\times d}$, $W_d\in\R^{d_o\times m}$,
$\sigma$ acts elementwise, and $\odot$ is the elementwise product. Biases are
omitted; fixed biases can be retained in the same identities.

\paragraph{Proposition A.1 (Realizing the History-Induced Input Change).}
\label{prop:app-mlp}
Keep $W_d$ fixed and set
\begin{equation}
  \Delta W_g=W_g\widehat P_H,
  \qquad
  \Delta W_u=W_u\widehat P_H.
  \label{eq:app-gate-up-update}
\end{equation}
Then both updates have rank at most $r$, and the updated MLP satisfies
\begin{equation}
  F_\Delta(x)=F\bigl((I+\widehat P_H)x\bigr)
  \quad\text{for every }x.
  \label{eq:app-mlp-identity}
\end{equation}
\paragraph{Proof.}
The updated gate preactivation is
$(W_g+\Delta W_g)x=W_g(I+\widehat P_H)x$; the same identity holds for the up
projection. Substituting both expressions into Eq.~\eqref{eq:app-gated-mlp} proves
Eq.~\eqref{eq:app-mlp-identity}. The explicit factorizations are
\begin{equation}
\begin{aligned}
  \Delta W_g&=(W_gW_OB_H)(A_HW_Q),\\
  \Delta W_u&=(W_uW_OB_H)(A_HW_Q).
\end{aligned}
  \label{eq:app-lora-factors}
\end{equation}
Each product has an inner dimension of $r$, so each update has rank at most $r$.
\hfill$\square$

Equation~\eqref{eq:app-lora-factors} has the form used by LoRA
\citep{hu2021lora}; any fixed nonzero adapter scale can be absorbed into a factor. The MLP
nonlinearity is preserved exactly because the update changes its inputs before
the activation and multiplication. In this construction, gate and up updates
work together to apply the same history-induced input change.

\paragraph{How Read Error Reaches the MLP Output.}
Write $E_H=M_H-\widehat M_H$. The feature error is
\begin{equation}
  x_H-(I+\widehat P_H)x=W_OE_HW_Qx.
  \label{eq:app-feature-error}
\end{equation}
If $F$ is $L_F$-Lipschitz on the segment between these two inputs, then
\begin{equation}
  \norm{y_H-F_\Delta(x)}_2
  \leq L_F\norm{W_O}_2\norm{E_HW_Qx}_2.
  \label{eq:app-output-error}
\end{equation}
Here matrix norms are spectral norms. For a truncated rank-$r$ singular value
decomposition, $\norm{E_H}_2=\sigma_{r+1}(M_H)$, and therefore
\begin{equation}
  \norm{y_H-F_\Delta(x)}_2
  \leq L_F\norm{W_O}_2\,
  \sigma_{r+1}(M_H)\norm{W_Qx}_2.
  \label{eq:app-svd-error}
\end{equation}
When $r\geq\rank(M_H)$, the factorization is exact and the error is zero. For
smaller ranks, Eq.~\eqref{eq:app-output-error} ties the approximation to the queries
actually encountered, which can be more informative than a worst-case bound.

\paragraph{The Down Projection.}
The same target can also be expressed at the MLP's intermediate features. Define
\begin{equation}
  z(x)=\sigma(W_gx)\odot(W_ux),
  \qquad z_H(x)=z\bigl((I+P_H)x\bigr).
  \label{eq:app-intermediate-features}
\end{equation}
For a fixed history, suppose one shared matrix $S_H\in\R^{m\times m}$,
with rank at most $s$, approximates the change across current inputs:
\begin{equation}
  z_H(x)=(I+S_H)z(x)+e_H(x).
  \label{eq:app-intermediate-fit}
\end{equation}
Keeping gate and up fixed and setting $\Delta W_d=W_dS_H$ gives
\begin{equation}
\begin{aligned}
  F_{\Delta d}(x)&=(W_d+\Delta W_d)z(x),\\
  \norm{y_H-F_{\Delta d}(x)}_2
    &\leq\norm{W_d}_2\norm{e_H(x)}_2,
  \qquad \rank(\Delta W_d)\leq s.
\end{aligned}
  \label{eq:app-down-error}
\end{equation}
For $S_H=D_HC_H$, the update is $\Delta W_d=(W_dD_H)C_H$, another LoRA
factorization. A useful low-rank $S_H$ is an additional property of the
intermediate features to measure; the nonlinear map $z$ makes it a separate
condition from low rank of $M_H$.

The gate/up construction and the down construction describe two ways to
reproduce the same target $y_H$. Jointly trained adapters can distribute the
correction across these projections. \method learns module-specific factors; the
identities above exhibit computations available within that parameterization.

\subsection{Updating the Historical Effect Online}
\label{appendix:math:online-update}

When a new clip $C$ arrives, its key--value pairs add a new contribution:
\begin{equation}
  M_{H\cup C}=M_H+\Delta M_C,
  \qquad
  \Delta M_C=\sum_{i\in C}v_i k_i^\top.
  \label{eq:app-online-matrix}
\end{equation}
Here $H\cup C$ denotes appending the clip to the history. With fixed $W_Q$ and
$W_O$, the feature transformation follows the same update:
\begin{equation}
  P_{H\cup C}=P_H+W_O\Delta M_CW_Q.
  \label{eq:app-online-feature}
\end{equation}
The update therefore extends the same effect traced from the historical read
to the MLP input. Previously produced keys and values are assumed to stay
fixed; the matrix retains their summed outer-product contribution.

\paragraph{Cost and Rank.}
For $n$ tokens per clip, adding the outer products costs
$O(nd_vd_k)$ per clip and stores $d_vd_k$ matrix entries. Over $T$ clips,
incremental accumulation costs $O(Tnd_vd_k)$. Rebuilding the matrix from all
available key--value pairs at every step costs
$O(T^2nd_vd_k)$. These costs cover dense matrix accumulation given the keys and
values. Video encoding, compression, adapter generation, and responses add
their respective costs.

Each new clip can increase the matrix rank. Maintaining a fixed rank can be
written as an update followed by compression:
\begin{equation}
  \widehat M_t=\mathcal C_r
  \left(\widehat M_{t-1}+\Delta M_{C_t}\right),
  \label{eq:app-rank-update}
\end{equation}
where $\mathcal C_r$ returns a matrix of rank at most $r$. The compression
algorithm determines both its additional cost and which historical responses
the bounded-rank memory preserves.

\paragraph{Connection to \method.}
\method implements an online memory with a learned state update $U_\phi$ and
adapter generator $D_\phi$:
\begin{equation}
  \Delta\theta_t=D_\phi(h_{t-1}),\qquad
  h_t=U_\phi\bigl(h_{t-1},E(C_t)\bigr).
  \label{eq:app-icic-update}
\end{equation}
Here $E(C_t)$ encodes the incoming clip. The response at step $t$ uses
$\Delta\theta_t$ together with the current clip; writing that clip into $h_t$
prepares the memory for later responses.
Equations~\eqref{eq:app-online-matrix}--\eqref{eq:app-rank-update} describe an analytical
update-and-compression process. \method learns a surrogate from the task objective:
its recurrent state and generated adapters aim to preserve the historical
effects useful for later responses. Their behavior can be compared with the
matrix read and feature transformation derived above.

\subsection{Extensions to the Backbone Computation}
\label{appendix:math:backbone-extensions}

The same reasoning can include the remaining operations by retaining their
normalization terms or using a local affine approximation.

\paragraph{Normalized Kernel Reads.}
For a finite feature map $\phi:\R^{d_k}\to\R^p$ and weights
$\kappa(q,k)=\phi(k)^\top\phi(q)$, define
\begin{equation}
  N_H=\sum_{i\in H}v_i\phi(k_i)^\top,
  \qquad b_H=\sum_{i\in H}\phi(k_i).
  \label{eq:app-kernel-state}
\end{equation}
If the denominator is positive, the normalized read is exactly
\begin{equation}
  R_H(q)=\frac{N_H\phi(q)}{b_H^\top\phi(q)}.
  \label{eq:app-kernel-read}
\end{equation}
Both $N_H$ and $b_H$ update additively by clip. When a read includes current
tokens $C$, their contributions are combined before division:
\begin{equation}
  R_{H\cup C}(q)=
  \frac{(N_H+N_C)\phi(q)}{(b_H+b_C)^\top\phi(q)}.
  \label{eq:app-kernel-joint}
\end{equation}
If $N_H$ is approximated by $\widehat N_H$, $b_H$ is kept exact, and
$b_H^\top\phi(q)\geq\eta>0$, then
\begin{equation}
  \norm{R_H(q)-\widehat R_H(q)}_2
  \leq \frac{\norm{(N_H-\widehat N_H)\phi(q)}_2}{\eta}.
  \label{eq:app-kernel-error}
\end{equation}
The denominator state is therefore part of the historical read, alongside its
numerator matrix.

\paragraph{Softmax Reads.}
For fixed historical keys and values and temperature $\tau>0$, let
\begin{equation}
  R_H(q)=\sum_{i\in H}p_i(q)v_i,
  \qquad
  p_i(q)=\frac{\exp(k_i^\top q/\tau)}
                    {\sum_{j\in H}\exp(k_j^\top q/\tau)}.
  \label{eq:app-softmax-read}
\end{equation}
Its local dependence on a query is described by an offset and a Jacobian.
Writing $\mu_k(q)=\sum_i p_i(q)k_i$, direct differentiation gives
\begin{equation}
  J_H(q)=\frac{1}{\tau}\sum_{i\in H}p_i(q)
       \bigl(v_i-R_H(q)\bigr)\bigl(k_i-\mu_k(q)\bigr)^\top.
  \label{eq:app-softmax-jacobian}
\end{equation}
Around a reference query $q_0$, with $\delta=q-q_0$,
\begin{equation}
  R_H(q)=R_H(q_0)+J_H(q_0)\delta+\rho_H(q).
  \label{eq:app-softmax-affine}
\end{equation}
If $J_H$ is $L_J$-Lipschitz along the segment from $q_0$ to $q$, the integral
form of the Taylor remainder gives
\[
  \norm{\rho_H(q)}_2\leq\tfrac12L_J\norm{\delta}_2^2.
\]
For a rank-$r$ matrix $\widehat J_H$ satisfying
$\norm{J_H(q_0)-\widehat J_H}_2\leq\epsilon_r$, define the affine read as
$\widehat R_H(q)=R_H(q_0)+\widehat J_H\delta$. It satisfies
\begin{equation}
  \norm{R_H(q)-\widehat R_H(q)}_2
  \leq\epsilon_r\norm{\delta}_2
      +\tfrac12L_J\norm{\delta}_2^2.
  \label{eq:app-softmax-error}
\end{equation}
This gives a local low-rank interpretation with the reference response
$R_H(q_0)$ retained explicitly.

When softmax attention reads both history and current tokens, define
\[
  Z_H(q)=\sum_{i\in H}\exp(k_i^\top q/\tau),
  \qquad
  Z_C(q)=\sum_{i\in C}\exp(k_i^\top q/\tau).
\]
For nonempty $H$ and $C$,
\begin{equation}
  R_{H\cup C}(q)=w_H(q)R_H(q)
                 +\bigl(1-w_H(q)\bigr)R_C(q),
  \quad w_H(q)=\frac{Z_H(q)}{Z_H(q)+Z_C(q)}.
  \label{eq:app-softmax-mixture}
\end{equation}
Thus adding history changes both the available values and their relative
weights. A reconstruction of the joint read must retain this mixing weight
together with the historical response.

\paragraph{Normalization and the MLP Input.}
In a pre-normalized backbone, the query and MLP each receive normalized
features. A controlled comparison can isolate the historical read within one
block. Fix the key--value sets, and let both paths receive the same residual
feature $u$. With attention input normalization $N_a$, MLP input normalization
$N_m$, and $q(u)=W_QN_a(u)$, the two MLP inputs are
\begin{equation}
\begin{aligned}
  T_H(u)&=N_m\left(u+W_OR_{H\cup C}(q(u))\right),\\
  T_0(u)&=N_m\left(u+W_OR_C(q(u))\right).
\end{aligned}
  \label{eq:app-normalized-interface}
\end{equation}
These expressions place both normalization operations and softmax mixing
inside the feature map being compared. For
$D_H(u)=T_H(u)-T_0(u)$, a local expansion at $u_0$ is
\begin{equation}
  D_H(u)=D_H(u_0)+J_D(u_0)(u-u_0)+r_D(u).
  \label{eq:app-normalized-affine}
\end{equation}
The constant term records the historical change at the reference feature; the
Jacobian records how that change varies nearby. If $J_D$ is $L_D$-Lipschitz on
the segment, $\norm{r_D(u)}_2\leq\tfrac12L_D\norm{u-u_0}_2^2$.

\paragraph{Comparing Complete Forward Passes.}
Removing history can also change earlier layers'
outputs and the current-token keys and values. Let $x_H^{\mathrm{act}}$ and
$x_0^{\mathrm{act}}$ be the normalized MLP inputs for the same current token
in retained-history and current-only runs of the frozen backbone. For a fixed
history, fit one rank-$r$ matrix $P$ across these pairs and measure the residual:
\begin{equation}
  x_H^{\mathrm{act}}=(I+P)x_0^{\mathrm{act}}+e_H^{\mathrm{in}}.
  \label{eq:app-backbone-input-fit}
\end{equation}
The gate/up construction with $\Delta W_g=W_gP$ and $\Delta W_u=W_uP$ gives
\begin{equation}
  \norm{F(x_H^{\mathrm{act}})-F_\Delta(x_0^{\mathrm{act}})}_2
  \leq L_F\norm{e_H^{\mathrm{in}}}_2,
  \label{eq:app-backbone-output-error}
\end{equation}
provided $F$ is $L_F$-Lipschitz on the segment from $x_H^{\mathrm{act}}$ to
$(I+P)x_0^{\mathrm{act}}$. The residual
measures how well the shared $P$ captures the full history-induced input
change, including normalization and changes from earlier layers. This connects
the block interpretation to a measurable approximation in the backbone.

In conclusion, this analysis establishes the theoretical feasibility of compressing the effect of historical context into low-rank parameter updates of the MLP modules within a Transformer block. In particular, it shows that, under the assumptions described above, a history-dependent feature transformation induced by historical context can be realized or approximated through LoRA-style updates to the MLP projections. Importantly, this analysis does not claim that \method exactly reconstructs the analytical history matrix \(M_H\), nor that its learned recurrent state and generated adapters are equivalent to full-context attention. Rather, the derivation provides a mathematical interpretation demonstrating that such a parameter-space representation of historical information is feasible. Our empirical results further show that \method provides an effective learned instantiation of this approximation in practice.

\begin{table*}[t]
\centering
\renewcommand{\arraystretch}{1.1}
\setlength{\tabcolsep}{5pt}
\caption{Distribution of training videos across underlying datasets and data sources.}
\vspace{-0.5em}
\label{tab:training_video_distribution}
\medskip
\resizebox{\textwidth}{!}{%
\begin{tabular}{lrrrrrr}
\toprule
\textbf{Underlying source} &
\textbf{TimeLens} &
\textbf{LLaVA-Video} &
\textbf{ET-Instruct} &
\textbf{Others} &
\textbf{Unique Wideos} &
\textbf{Weight} \\
\midrule
CosMo-Cap~\citep{wang2024cosmo}
    & 9,524 & --    & --    & --    & \textbf{9,524} & \textbf{17.05\%} \\
YouCook2~\citep{ZhXuCoAAAI18}
    & --    & 7,451 & --    & 1,704 & \textbf{9,155} & \textbf{16.39\%} \\
Charades~\citep{gao2017tall}
    & --    & 8,692 & --    & --    & \textbf{8,692} & \textbf{15.56\%} \\
ActivityNet~\citep{krishna2017dense}
    & --    & 7,543 & --    & --    & \textbf{7,543} & \textbf{13.50\%} \\
InternVid-VTime~\citep{wang2024internvid}
    & 4,644 & --    & --    & --    & \textbf{4,644} & \textbf{8.31\%} \\
NextQA~\citep{xiao2021next}
    & --    & 3,704 & --    & --    & \textbf{3,704} & \textbf{6.63\%} \\
DiDeMo~\citep{anne2017localizing}
    & 2,948 & --    & --    & --    & \textbf{2,948} & \textbf{5.28\%} \\
STAR~\citep{wu2024star}
    & --    & 2,008 & --    & --    & \textbf{2,008} & \textbf{3.59\%} \\
COIN~\citep{tang2019coin}
    & --    & --    & 1,995 & --    & \textbf{1,995} & \textbf{3.57\%} \\
CrossTask~\citep{zhukov2019cross}
    & --    & --    & --    & 1,640 & \textbf{1,640} & \textbf{2.94\%} \\
QuerYD~\citep{oncescu2021queryd}
    & 1,511 & --    & --    & --    & \textbf{1,511} & \textbf{2.70\%} \\
HiREST~\citep{zala2023hierarchical}
    & 1,204   & --    & --    & --   & \textbf{1,204} & \textbf{2.16\%} \\
Ego4D~\citep{grauman2022ego4d}
    & --    & 905   & --    & --    & \textbf{905}   & \textbf{1.62\%} \\
THUMOS14~\citep{idrees2017thumos}
    & --    & --    & --    & 394   & \textbf{394}   & \textbf{0.71\%} \\
\midrule
\textbf{Total} &
\textbf{19,831} &
\textbf{30,303} &
\textbf{1,995} &
\textbf{4,155} &
\textbf{55,867} &
\textbf{100.00\%} \\
\bottomrule
\end{tabular}%
}
\end{table*}

\section{Additional Dataset Details} \label{appendix:data}
In this section, we provide more information related to the preparation of \method-Instruct. We share the video source details in section~\ref{appendix:data:video_source}, the data pipeline and the annotation construction details in section~\ref{appendix:data:data_pipeline}, and finally the dataset statistics in section~\ref{appendix:data:statistic}.

\subsection{Video Sources} \label{appendix:data:video_source}
We begin the dataset construction by collecting videos from multiple existing sources, including TimeLens-100K~\citep{zhang2026timelens}, the academic subset of LLaVA-Video-178K~\citep{zhang2024llava}, ET-Instruct~\citep{liu2024bench}, and several additional video datasets, including YouCook2~\citep{ZhXuCoAAAI18}, CrossTask~\citep{zhukov2019cross}, and THUMOS14~\citep{idrees2017thumos}. We primarily build upon these large-scale collections rather than gathering every underlying dataset independently, as they already consolidate videos from diverse academic sources and have undergone prior curation and preprocessing, providing a more scalable and reliable starting point for dataset construction. To prevent evaluation contamination, we further filter out videos that overlap with those used in OVO-Bench~\citep{niu2025ovo}. In contrast, StreamingBench and StreamArena do not overlap with any of our video sources and therefore require no additional filtering. After aggregating all sources and removing duplicate and overlapping videos, the resulting dataset contains 55,867 unique videos drawn from 14 underlying video datasets. A detailed breakdown of the video distribution is provided in Table~\ref{tab:training_video_distribution}.

\subsection{Data Pipeline and Annotation Construction} \label{appendix:data:data_pipeline}
As illustrated in Figure~\ref{fig:data_pipeline}-1, our data construction pipeline consists of three components: a \textit{Question Generator}, an \textit{Answer Generator}, and a \textit{Verifier}. To construct the streaming video training dataset, we begin with an event description associated with a specific temporal interval. These descriptions are obtained either from the original annotations of the 14 datasets described in Section~\ref{appendix:data:video_source}, or from the question--answer pairs provided by TimeLens and ET-Instruct.

\textbf{Question Generation.}
We first prompt the \textit{Question Generator} to generate a new question for a target event, conditioned on the full video, the event description, and its corresponding temporal interval. By observing the full video, the \textit{Question Generator} can leverage both the target event and its surrounding visual context to construct a question that is grounded in the video content. The prompt explicitly controls the question type, which is sampled from four categories: open-ended, multiple-choice, Yes/No, and counting questions. Along with each generated question, the \textit{Question Generator} also produces an \textit{expected answer}, which is subsequently used during the verification stage.
To cover different temporal reasoning scenarios, we further assign each question a randomly sampled question-timestamp type from three categories: \textit{past}, \textit{real-time}, and \textit{future}. Given a target event spanning the interval $[t_{\mathrm{start}}, t_{\mathrm{end}}]$, a \textit{past} question is assigned a timestamp $t > t_{\mathrm{end}}$, requiring the model to retrospectively recall an event that has already occurred. A \textit{real-time} question is assigned a timestamp $t \in [t_{\mathrm{start}}, t_{\mathrm{end}}]$, such that the target event is ongoing when the question is asked. A \textit{future} question is assigned a timestamp $t < t_{\mathrm{start}}$, requiring the model either to anticipate a future event from the available context or to abstain from answering when the observed evidence is insufficient. The exact question timestamp is then randomly sampled from the temporal range corresponding to the selected timestamp type.
We use Qwen3.5-122B-A10B~\citep{qwen3.5} as the \textit{Question Generator}. For question generation, we launch four parallel workers, each serving one Qwen3.5-122B-A10B instance with vLLM~\citep{kwon2023efficient} on two H200 GPUs.

\textbf{Answer Generation.}
After obtaining the generated questions and their corresponding timestamps $t$, we prompt the \textit{Answer Generator} to produce a \textit{model answer} using only the video content observable up to time $t$. This setup ensures that the generated answer respects the streaming setting and does not access visual information occurring after the question timestamp. We use Qwen3.5-122B-A10B~\citep{qwen3.5} as the \textit{Answer Generator}. For answer generation, we launch four parallel workers, each serving one Qwen3.5-122B-A10B instance with vLLM on two H200 GPUs.

\textbf{Verification.}
After obtaining the generated question and its corresponding \textit{model answer}, we prompt the \textit{Verifier} to assess the quality of the generated question--answer pair. The \textit{Verifier} takes as input the generated question, the \textit{expected answer}, the \textit{model answer}, and the original event description used during question generation. It evaluates both whether the generated question is valid and well grounded in the target event and whether the \textit{model answer} is consistent with the available evidence. In particular, a substantial mismatch between the \textit{expected answer} and the \textit{model answer} may indicate either that the generated question is ambiguous or unreasonable, or that the \textit{model answer} is incorrect. The original event description provides additional grounding information that allows the \textit{Verifier} to distinguish between these cases and assess the overall validity of the question--answer pair. For verification, we launch two parallel workers, each serving one Qwen3.5-397B-A17B instance with vLLM on four H200 GPUs.

\textbf{Filtering and Post-Processing.}
We discard question--answer pairs for which the \textit{Verifier} identifies an inconsistency between the \textit{expected answer} and the \textit{model answer}, determines that the \textit{model answer} is incorrect, or judges the generated question to be invalid, ambiguous, or otherwise unreasonable. After filtering, we group questions associated with the same video and question timestamp into multi-turn conversations, which are subsequently used for training.
We also include annotations from the offline video understanding from LLaVA-Video-178K to further strengthen the memory mechanism.

\subsection{Dataset Statistic} \label{appendix:data:statistic}
\method-Instruct contains 381K instruction samples constructed from 56K videos.
Table~\ref{tab:timestamp_distribution} and Table~\ref{tab:question_type_distribution} summarize the distribution of questions by timestamp class and question type. In terms of temporal categories, real-time questions constitute the largest proportion (46.1\%), followed by past (36.4\%) and future (17.4\%) questions. These categories correspond to questions posed while the target event is occurring, after the event has occurred, and before the event occurs, respectively. Regarding question types, multiple-choice questions account for the largest proportion (35.0\%), while open-ended (34.7\%) and Yes/No (30.0\%) questions are comparably represented, with counting questions comprising the remaining 14.0\%.

\begin{figure*}[t]
    \centering
    \includegraphics[width=0.99\linewidth]{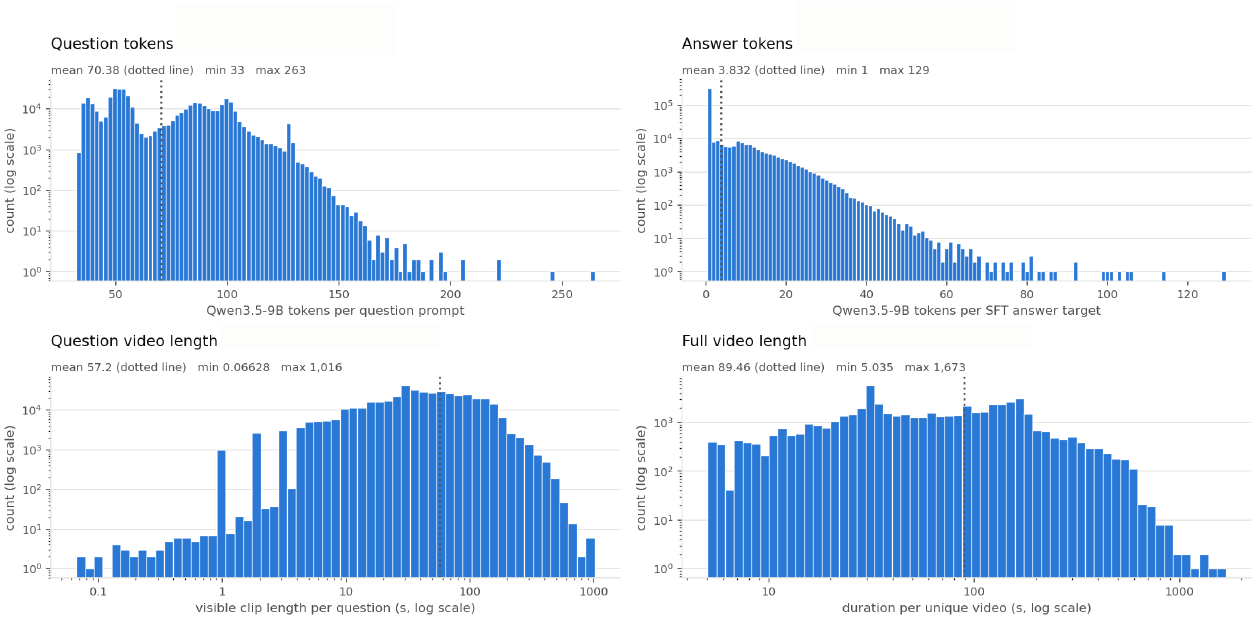}
    \vspace{-0.5em}
    \caption{The distribution of questions, answers, and video durations in the dataset.}
    \label{fig:appendix_data_distri}
    \vspace{-0.5em}
\end{figure*}

Table~\ref{tab:dataset_statistics} summarizes the statistics of the questions, answers, and video durations in the dataset. Questions contain an average of 70.4 tokens, encoded by the tokenizer from Qwen3.5-9B, while answers are generally concise, with an average of 3.8 tokens. The question timestamps cover a wide temporal range, with an average question video length of 57.2 seconds, while the full videos are longer on average at 89.5 seconds, reaching up to 1,673.4 seconds. The distribution of questions, answers, and video durations in the dataset are presented in Figure~\ref{fig:appendix_data_distri}.

\begin{table*}[t]
    \centering

    \begin{minipage}[t]{0.48\textwidth}
        \vspace{0pt}
        \centering
        \caption{Distribution of questions across timestamp classes.}
        \vspace{-0.5em}
        \label{tab:timestamp_distribution}
        \medskip
        \scalebox{0.88}{
        \begin{tabular}{lrr}
            \toprule
            \textbf{Timestamp class} & \textbf{Questions} & \textbf{Raw corpus share} \\
            \midrule
            Realtime & 175,773 & 46.1\% \\
            Past     & 138,600 & 36.4\% \\
            Future   & 66,547  & 17.4\% \\
            \midrule
            \textbf{Total} & \textbf{380,920} & \textbf{100.00\%} \\
            \bottomrule
        \end{tabular}
        }
    \end{minipage}
    \hfill
    \begin{minipage}[t]{0.48\textwidth}
        \vspace{0pt}
        \centering
        \caption{Distribution of questions across question types.}
        \vspace{-0.5em}
        \label{tab:question_type_distribution}
        \medskip
        \scalebox{0.88}{
        \begin{tabular}{lrr}
            \toprule
            \textbf{Question type} & \textbf{Questions} & \textbf{Raw corpus share} \\
            \midrule
            Multiple-choice & 120,416 & 35.0\% \\
            Open-ended      & 119,371 & 34.7\% \\
            Yes/no          & 104,508 & 30.0\% \\
            Counting        & 36,625  & 14.0\% \\
            \midrule
            \textbf{Total} & \textbf{380,920} & \textbf{100.00\%} \\
            \bottomrule
        \end{tabular}
        }
    \end{minipage}

\end{table*}

\begin{table}[t]
\centering
\renewcommand{\arraystretch}{1.1}
\setlength{\tabcolsep}{8pt}
\vspace{-0.5em}
\caption{Statistics of questions, answers, and video durations in the dataset.}
\vspace{-0.5em}
\label{tab:dataset_statistics}
\medskip
\scalebox{0.88}{
\begin{tabular}{lrrr}
\toprule
& \textbf{Min} & \textbf{Max}  & \textbf{Mean} \\
\midrule
Question tokens
& 33 & 263  & 70.4 \\

Answer tokens
& 1 & 129  & 3.8 \\

Question timestamp (s)
& 0.07 & 1,015.8  & 57.2 \\

Full video length (s) & 5.04 & 1,673.4  & 89.5 \\
\bottomrule
\end{tabular}
}
\vspace{-1em}
\end{table}

\section{Additional Experiment Details}\label{appendix:additional_details}

\subsection{Main Experiments} \label{appendix:additional_details:main}
In this section, we provide additional experiment details for the main experiments presented in Table~\ref{table:main_result}.

\textbf{Base Models.} In the main experiments, we build \method on top of four different models, VideoChat3-4B~\citep{li2026videochat3}, Qwen3-VL-4B~\citep{bai2025qwen3}, Qwen3.5-9B~\citep{qwen3.5}, and MOSS-VL-Realtime-11B~\citep{wang2026moss}. 
VideoChat3 is a video-centric VLM that is further fine-tuned on streaming video understanding data and adopts a full self-attention architecture.
Qwen3-VL-4B is a general-purpose VLM based on a full self-attention architecture.
Qwen3.5 is a general-purpose VLM based on a hybrid attention architecture, where a self-attention layer is interleaved with every four linear-attention layers.
MOSS-VL is a video-centric VLM that adapts cross-attention layers to inject visual information into the text tokens, and a cross-attention layer is interleaved with every four self-attention layers.

\textbf{Video Preprocessing.} During training and evaluation, we sample the video frames at 1 fps, dividing the video frames into non-overlapping two-frame temporal blocks (each $C_t$ is a 2-second video clip). We regard the most recent 2-second block as the current observation, which will be sent into the VLMs, while all the rest of the history frames will be sent into the \method to be converted into LoRA weights. 
For \method(Qwen3-VL-4B), we observe that using longer current-observation windows consistently improves performance on the real-time and backward splits of OVO-Bench~\citep{niu2025ovo} and StreamingBench~\citep{lin2026streamingbench}. We therefore use the most recent 8-second clips as the current observations for \method(Qwen3-VL-4B), while the other \method instantiations in the table use the most recent 2 seconds. We hypothesize that these findings, \method(Qwen3-VL-4B) benefiting from longer current-observation windows and the particularly strong performance on StreamingBench, are largely attributable to differences in the architecture and pretraining recipe of the underlying VLMs.
We preprocess the video frames with the preprocessor from the corresponding VLMs, with the maximum per-frame pixel budget set to 262,144 for all three versions of \method.

\textbf{Vision Encoder.} We use the vision encoder from the corresponding VLMs as the vision encoder to encode each of the temporal blocks into vision tokens. We also use the MLP layers or adaptor from the VLMs to convert the vision tokens into the text semantic space before we send them into the Aggregator. We denote the vision tokens from the temporal block at $t$ as $V_t$.

\textbf{Aggregator.} Visual tokens $V_t$ in the temporal block at $t$ are concatenated with the learnable queries $Q$ and the corresponding text embedding $T_t$ of the timestamp of the temporal block. These tokens are first projected to a hidden dimension of 512 and processed by a four-layer Transformer encoder aggregator with eight attention heads (64 dimensions per head), a feed-forward dimension of 2,048, pre-layer normalization, GELU activations. We learn one 512-dimensional query for each of the VLM decoder layers. The query is shared across target modules, and LoRA ranks in the same VLM decoder layer.

\textbf{State-Space Model.}
We learn a four-layer DeltaNet~\citep{yang2024parallelizing} with a hidden dimension of 512 as the state-space model. Each DeltaNet layer uses four value heads and two key heads, both with head dimension 128. 

\textbf{Cosine-Similarity-Based Gating.}
We determine the gating threshold by profiling the cosine similarity score between two adjecent block from 1000 subset of video from the training data introduced in section~\ref{section:Training_data}. Across 64,426 transitions between consecutive temporal blocks, the median raw cosine distance,
$1-\cos(\bar{V}_t,\bar{V}_{t-1})$, is $0.0542$. We therefore set $\beta=0.05$, approximately the median-change operating point. The first temporal block is always retained, while each subsequent block is retained if $\cos(\bar{V}_t,\bar{V}_{t-1}) \leq \beta$.
Blocks exhibiting less visual change than this threshold receive zero weight in the residual-history aggregation, whereas the retained blocks are uniformly averaged. 

\textbf{The Converter.}
For each target-module type and each LoRA rank, a separate linear output head maps the 512-dimensional representation $Norm(o_{t}) + \alpha Norm(s_{t})$ to a vector of dimension $d_{\mathrm{in}}+d_{\mathrm{out}}$. The vector is split into one row of the LoRA factors $\mathbf{A}_{t}\in\mathbb{R}^{r\times d_{\mathrm{in}}}$ and $\mathbf{B}_{t}\in\mathbb{R}^{r\times d_{\mathrm{out}}}$. These output heads are shared across decoder layers, while layer specificity is supplied by the layer-specific learned queries. Finally, the LoRA at a particular slot $s$ at timestamp t is calculated with 
$\Delta \theta_t=\alpha\,B_t^{\mathsf T}A_t$, where \(\alpha\) is lora\_alpha, which we set to 16.

\textbf{LoRA Application Location.}
For VideoChat3-4B, we apply rank-8 generated LoRA for the MLP projections (\texttt{gate\_proj}, \texttt{up\_proj}, and \texttt{down\_proj}) and \texttt{self\_attn.v\_proj} in every attention layer.
For the Qwen3-VL-4B, we apply rank-8 generated LoRA for the MLP projections (\texttt{gate\_proj}, \texttt{up\_proj}, \texttt{down\_proj}), and \texttt{self\_attn.v\_proj} in every attention layer.
For Qwen3.5-9B, we apply rank-8 generated LoRA for the MLP projections (\texttt{gate\_proj}, \texttt{up\_proj}, and \texttt{down\_proj}) in every attention layer, \texttt{self\_attn.v\_proj} in full-attention layers, and \texttt{linear\_attn.in\_proj\_qkv} in linear-attention layers.
For MOSS-VL-Realtime-11B, we apply rank-8 generated LoRA for the MLP projections (\texttt{gate\_proj}, \texttt{up\_proj}, and \texttt{down\_proj}), and \texttt{v\_proj} in every attention layer. We inject the generated LoRA weights into the target linear layers of the frozen base model through customized forward functions.

\textbf{Training Details.}
We build our training pipeline on top of the Llama-Factory~\citep{zheng2024llamafactory}.
We train \method on the \method-Instruct dataset introduced in Section~\ref{section:Training_data} for 2,800 optimization steps with a total batch size of 64. Training uses four H200s, a per-device batch size of one, and 16 gradient-accumulation steps. We use AdamW with $\beta_1=0.9$, $\beta_2=0.999$, $\epsilon=10^{-8}$, an peak learning rate of $8\times10^{-5}$, a cosine learning-rate schedule, and a warm-up ratio of $0.1$. Training is performed in bfloat16. We set the LoRA regularization factor $\gamma$ to 0.01.

During training, we use the state-space model in its parallel causal form. For each sample, the compact vision representations $q_t$ are arranged in chronological order and processed by the four-layer DeltaNet using the chunk-parallel gated-delta-rule kernel and an attention mask for padded blocks. We then select the output at the last valid temporal block for each history observation. This parallel computation is causally equivalent to processing the blocks recurrently while carrying the convolutional and recurrent DeltaNet states, which is the form that is used for online streaming inference.

\textbf{Evaluation Details.}
Following the evaluation protocols adopted in prior streaming video understanding works~\citep{shen2026simple, zhang2026hermes}, we evaluate \method on the \textit{Real-Time} split of StreamingBench~\citep{lin2026streamingbench}, as well as the \textit{Real-Time} and \textit{Backward} splits of OVO-Bench~\citep{niu2025ovo}. 

For StreamingBench, we evaluate on the \textit{Real-Time} split, which contains 2,500 multiple-choice questions, and use accuracy as metric. For OVO-Bench, both the \textit{Backward} and \textit{Real-Time} splits are evaluated with the metric accuracy. The \textit{Backward} and \textit{Real-Time} splits consist of 631 and 837 questions. Following prior work, the performance of each split is computed as the macro-average across the tasks belonging to that split, and we additionally report the macro-average performance across the \textit{Backward} and \textit{Real-Time} splits in Table~\ref{table:main_result}.

The \textit{Backward} splits of OVO-Bench mostly focus on retrospect to the past and comprise three tasks: Episodic Memory (EPM), Action Sequence Identification (ASI), and Hallucination Detection (HLD). Among them, EPM is particularly well suited for evaluating the memory mechanism introduced by \method, as it directly measures the model's ability to retain and retrieve information from earlier observations. We further note that HLD is relatively less challenging because all ground-truth responses are \textit{Unable to answer}. As a result, models can achieve high accuracy on this task by consistently predicting this response, making HLD less discriminative for assessing the effectiveness of long-term memory mechanisms.

The trained \method can operate in a fully streaming manner by processing the video history block by block using the cached-forward mode of DeltaNet. (Please see the \textit{Evaluation details.} in the section~\ref{appendix:additional_details:vla}) To accelerate evaluation, however, we use the same parallel-forward mode employed during training when evaluating on OVO-Bench and StreamingBench. Specifically, for a question associated with a video prefix $[0,T]$, we treat the interval $[0,T-2]$ as historical context and sample it at 1 fps as input to the hypernetwork. The base VLM then receives the question together with two frames sampled from the current observation window $[T-2,T]$. For the evaluation on the OVO-bench, we follow the prompt example used in lmms-eval~\citep{zhang2024lmmsevalrealitycheckevaluation}, and for the evaluation on the StreamingBench we follow the prompt example used in the official repository.

\textbf{Baselines Re-Evaluation.}
Among the baselines in Table~\ref{table:main_result}, we re-evaluate MOSS-VL-Realtime-11B~\citep{wang2026moss}, VideoChat3-4B~\citep{li2026videochat3}, and Qwen3.5-9B~\citep{qwen3.5}, as these models are also used as the base VLMs for \method. For the remaining baselines, we report the results from their original papers.

For Qwen3.5-9B, we sample video frames at 1 fps with a maximum per-frame pixel budget of 262,144 and use greedy decoding for answer generation.

For VideoChat3-4B, we use the official evaluation code and follow the hyperparameters reported in the paper. On OVO-Bench, we sample frames at 2 fps and retain the 32 most recent frames, with a per-frame pixel budget of $448^2$ and a visual-token budget of 80K. On StreamingBench, we sample frames at 4 fps with at most 4,096 frames, using the same per-frame pixel budget of $448^2$ and visual-token budget of 80K.

For MOSS-VL-Realtime-11B, as neither the paper nor the released repository provides an official evaluation script, we evaluate the model on OVO-Bench and StreamingBench using frames sampled at 1 fps with a maximum per-frame pixel budget of 262,144.

These settings successfully reproduce the reported OVO-Bench performance of VideoChat3 and MOSS-VL-Realtime. However, our re-evaluated StreamingBench results are lower than those reported in the corresponding papers. We therefore include both the re-evaluated and paper-reported results in Table~\ref{table:main_result}, with the paper-reported results shown in \textcolor{gray}{gray}.

\subsection{\method in VLA setting} \label{appendix:additional_details:vla}
In this section, we provide more implementation details for the \method in the VLA setting. The results are presented in Table~\ref{tab:vla}.

\textbf{Base VLA Model and Initialization.}
Inspired by the previous work~\citep{zhang2026vlmvla}, we train a simple VLA model on top of Qwen3.5-2B, named as Q3.5-2B-MLP, as the starting point of the \method(Q3.5-2B-MLP).

The Q3.5-2B-MLP is given the current observation (an image) and text instruction and predicts a chunk of four seven-dimensional actions, corresponding to the current action and the following three actions.
We learn special tokens as action hidden representations, and VLMs compress the visual observation and the text instruction into the special tokens.
An action decoder, which is a two-layer MLP that directly regresses normalized robot actions using mean-squared error based on the special tokens.
We use the Qwen3.5 preprocessor to preprocess the input image and the text instruction.

Q3.5-2B-MLP was trained on the Bridge and RT-1 mixture of the OXE LeRobot dataset~\citep{o2024open} for 50,000 optimization steps, with learning rate is \(4\times10^{-5}\) on action decoder, and \(1\times10^{-4}\) on LoRA. The learning rate is linearly warmed up for 5,000 steps and then cosine-decayed. 
Training uses AdamW with \(\beta_1=0.9\), \(\beta_2=0.95\), \(\epsilon=10^{-8}\),and weight decay \(10^{-8}\).
Training is conducted in bfloat16 on four H200 GPUs with a per-device batch size of 16 and two gradient-accumulation steps, yielding a global batch size of 128.
We apply rank-128 LoRA to all linear layers, including the vision tower and multimodal projector, with a scaling factor of 256. Q3.5-2B-MLP achieves 59.4\% in the Simpler-Bridge with WidowX robot.

\textbf{Observation Preprocessing.}
For a training sample at control step \(t\), the base VLA Model receives the current observation \(x_t\) and the language instruction, while the \method receives all preceding observations \(x_0,\ldots,x_{t-1}\).
We do not impose a maximum history length.
Bridge trajectories are recorded at 5 Hz, and timestamps are therefore expressed using a control period of \(0.2\) seconds.

The current observation is processed through the same single-image preprocessing path as the Q3.5-2B-MLP, while all the historical observations are decoded directly from the episode video and be sent into the \method. All the frames are resized to \(224\times224\), and passed to the Qwen3.5-2B preprocessor. This matches the preprocessing of the historical frames received from SimplerEnv during interactive evaluation.

We divide the observation history into consecutive, non-overlapping two-frame blocks, matching the temporal patch size of the Qwen vision encoder.
If the history contains an odd number of frames, its final frame is repeated to complete the last block.
For example, the history at step \(t=3\) is represented as
\([x_0,x_1]\) and \([x_2,x_2]\).
Each block is assigned a timestamp interval using the true Bridge control clock.
A committed block \(j\) is labeled
\([2j\Delta t,(2j+2)\Delta t]\), where \(\Delta t=0.2\) seconds.

\textbf{Vision Encoder.}
We use a frozen copy of the vision tower and vision adaptor from the Q3.5-2B-MLP as the \method's vision encoder.

\textbf{Action-History Conditioning.}
In addition to visual history, every historical frame is paired with the action that produced it.
Specifically, frame \(x_i\) is paired with the preceding action \(a_{i-1}\).
The first frame \(x_0\) is paired with a learned empty-action embedding representing \(a_{-1}\).
Historical actions undergo the same normalization transformations as the action targets.

Each seven-dimensional action is mapped to the visual-token dimension by a two-layer MLP with a GELU activation.
Thus, the input sequence for a two-frame temporal block is
\[
  [V_{2j},V_{2j+1},T_j,E(a_{2j-1}),E(a_{2j})],
\]
where \(V_i\) denotes the visual tokens for frame \(i\), \(T_j\) denotes the frozen token embeddings of the block's timestamp interval, and \(E(\cdot)\) is the action-history MLP.

\textbf{Aggregator.}
For every temporal block, the visual, timestamp, and action-history tokens are projected to a hidden dimension of 512 and processed by a four-layer Transformer encoder.
The rest of the aggregator's details are the same as the one used in section~\ref{appendix:additional_details:main}.

\textbf{State-Space Model and the Converter.} The state-space model and the converter share the same details as the one used in section~\ref{appendix:additional_details:main}.

\textbf{LoRA Application Locations.}
Because Qwen3.5 interleaves full self-attention and linear-attention layers,
the rank 8 generated LoRA is applied to
\texttt{gate\_proj}, \texttt{up\_proj}, and \texttt{down\_proj} in every decoder layer,
to \texttt{self\_attn.v\_proj} in full-attention layers, and to
\texttt{linear\_attn.in\_proj\_qkv} in linear-attention layers.

\textbf{Training Details.}
\method(Q3.5-2B-MLP) is trained on the Bridge portion of the OXE LeRobot dataset.
Before training, we merge the Q3.5-2B-MLP LoRA adapter into Qwen3.5-2B and load the stage-1 action-decoder weights.
The merged Qwen backbone is then frozen.
The trainable components in the \method(Q3.5-2B-MLP) are the \method, including the aggregator, action-history encoder, temporal state-space model, and LoRA output heads, together with the action decoder.
The complete objective is
\[
  \mathcal{L}
  =
  \mathcal{L}_{\mathrm{action}}
  + 0.1\,\mathcal{L}_{\mathrm{LoRA}},
\]
where
\[
  \mathcal{L}_{\mathrm{LoRA}}
  =
  \frac{1}{|\mathcal{S}|}
  \sum_{s\in\mathcal{S}}
  \left(
    \operatorname{mean}|\mathbf{A}_s|
    +
    \operatorname{mean}|\mathbf{B}_s|
  \right)
\]
, \(\mathcal{S}\) is the set of valid generated LoRA slots, and $\mathcal{L}_{\mathrm{action}}$ is MSE Loss between the predicted actions and the ground truth actions.

We train for 25,000 optimization steps using four H200 GPUs.
The per-device batch size is four and gradients are accumulated for four steps, giving a total batch size of 64.
We use AdamW with
\(\beta_1=0.9\), \(\beta_2=0.95\), \(\epsilon=10^{-8}\), and weight decay \(10^{-8}\) and learning rate of \(1\times10^{-5}\) in this experiment.
We use a cosine schedule with 1,500 warm-up steps and a final learning rate of \(5\times10^{-7}\).
Training is performed in bfloat16 with FlashAttention-2 and ordinary distributed data parallelism.

\textbf{Evaluation Details.}
We evaluate \method(Q3.5-2B-MLP) on the four WidowX Bridge tasks in SimplerEnv:
Stack Green Block on Yellow Block, Put Carrot on Plate, Put Spoon on Towel, and Put Eggplant in Yellow Basket.
Each task contains 24 evaluation episodes, for a total of 96 episodes.

Evaluation is conducted in a fully interactive streaming setting. At environment step zero, no historical context is available, and the model exactly reproduces the Q3.5-2B-MLP policy. At each subsequent step, completed two-frame blocks and the corresponding actions are permanently incorporated into the cached DeltaNet state $m_t$ and updated into the global sematic $s_t$. If the history ends with an unpaired frame, we duplicate it to form a provisional block and generate the current adapter using a temporary copy of the recurrent state. At the next step, we discard this temporary update and incorporate the completed two-frame block into the persistent state. The $h_t$ state is reset at each episode boundary.

The policy predicts four-action chunks and applies action ensembling over a horizon of four steps. Because ensembling may cause the action executed by the environment to differ from the most recent raw model prediction, we feed the normalized executed action back to the policy server. This executed action, rather than the unensembled model prediction, is paired with the corresponding observation when constructing subsequent history blocks.

\subsection{\method with different VLM scales and sources} \label{appendix:additional_details:mllms_scale_and_sources}
In this section, we provide additional implementation details for the results presented in Table~\ref{tab:different_scale_diffent_source}.

\textbf{Base Models.} In this sections, we use the Gemma-E4B~\citep{team2026gemma} and Qwen3.5-0.8B, Qwen3.5-2B and Qwen3.5-4B~\citep{qwen3.5} as the base model.

\textbf{LoRA Application Location.} For Gemma-E4B, we apply rank-8 generated LoRA for the MLP projections (\texttt{gate\_proj}, \texttt{up\_proj}, and \texttt{down\_proj}) in every attention layer.
For Qwen3.5 series, we apply rank-8 generated LoRA for the MLP projections (\texttt{gate\_proj}, \texttt{up\_proj}, and \texttt{down\_proj}) in every attention layer, \texttt{self\_attn.v\_proj} in full-attention layers, and \texttt{linear\_attn.in\_proj\_qkv} in linear-attention layers.

\textbf{Data, Structural, and Training Details.} For the Video Preprocessing, Vision Encoder, Aggregator, State-Space Model, Cosine-similarity-based gating, the Converter, and Training details are share the same details in section~\ref{appendix:additional_details:main}.

\textbf{Evaluation Details.}
For the evaluation on the OVO-bench, we follow the example used in lmms-eval~\citep{zhang2024lmmsevalrealitycheckevaluation}, and for the evaluation on the StreamingBench we follow the example used in the official repository.
For a question associated with video prefix $[0,T]$, the interval $[0,T-2]$ is treated as historical context and sampled at 1 fps for the \method. The base VLM receives the question together with two frames sampled from the current observation $[T-2,T]$. 

For OVO-Bench, we evaluate all backward, real-time, and forward splits. Backward and real-time questions are scored by accuracy. For the forward split, repetitive-event counting is scored by exact count matching, while sequential-step recognition and contextual-readiness recognition are scored using exact accuracy. We follow the benchmark's macro-averaging protocol across tasks for each split, and we further present the macro-averaging number over 3 splits.

For StreamingBench, we evaluate the Real, Omni, Sequential-QA, and Proactive subsets base on how the benchmark annotations are organized. Real, Omni, and Sequential-QA are scored by accuracy. Following the original protocol, Sequential-QA includes the gold question--answer history from earlier timestamps in the same video. For Proactive evaluation, the model is queried once per second until four seconds after the annotated trigger time. An example is counted as correct only when the model triggers within this tolerance window, and its generated response contains the expected output. The overall StreamingBench score is the benchmark-weighted average over Real, Omni, Proactive, and Sequential-QA, using weights of 2,500, 1,500, 250, and 250 examples, respectively.

\subsection{Memory stress test}\label{appendix:additional_details:memory_stress_test}
In this section, we share additional implementation details for the memory stress test evaluation and the results presented in Table~\ref{tab:StreamArena_hr}.

We use the StreamArena~\citep{zhang2026streamarena} Historical Retrospection (HR) split as the benchmark, which has 877 questions across L1–L4 over 237 videos. L1 to L4 indicate the evidence-to-query gap of less than 5 min, between 5 and 15 min, between 15 and 30 min, and longer than 30 min, respectively.

We compare \method (Qwen3.5-9B) against strong streaming video understanding models: MOSS-VL-Realtime-11B~\citep{wang2026moss}, which retains the full historical context; VideoChat3-4B~\citep{li2026videochat3} and AURA-8B~\citep{lu2026aura}, which operates with a local 16 / 30 -second observation window, respectively; VST-7B~\citep{guan2026video}, which compresses the historical observation as text-based memory.

For AURA and VST, we follow the official implementations in StreamArena to evaluate the model performance. For MOSS-VL-Realtime, VideoChat3, and \method, they share the same question prompt derived from the prompt used by AURA. 
\method (Qwen3.5-9B) samples video frames at 1fps with a maximum per-frame pixel budget of 262,144. The final two-second window is regarded as the current observation and will be sent to the VLMs, while the rest of the history frames will be sent to the \method and converted into LoRA. The VideoChat3 uses 32 unique frames at 2 FPS from the latest 16 seconds, with an aspect-ratio-preserving maximum of 200,704 pixels per frame (\(448^2\)). MOSS-VL samples frames at 1 FPS and with maximum per-frame pixel cap as 262,144 pixels. VST uses up to 384 causal-prefix frames at 1 FPS with aspect-ratio-preserving resolution determined dynamically by its 8,192-visual-token-per-segment budget. 
AURA takes most recent 30 seconds video at 1 FPS as input, with per-frame max pixel 602,112 pixel.

\subsection{Streaming Inference Speed Profiling}\label{appendix:additional_details:speed}
We provide additional details on the inference-speed profiling reported in Figure~\ref{fig:speed}. We measure the streaming inference latency of \method and four baselines on 100 deterministically sampled OVO-Bench~\citep{niu2025ovo} questions spanning a wide range of history lengths. Video durations range from 8 to 1,117 seconds, with a median of 153.5 seconds, a mean of 193.3 seconds, and a 90th percentile of 312 seconds. This set includes both short clips and videos exceeding 10 minutes, enabling evaluation of long-context streaming efficiency. At each query timestamp \(t\), we measure synchronized GPU computation required to process the current observation and answer the query under each method's native streaming protocol. Video decoding, resizing, and other input preprocessing are excluded.

For \method, each streaming step includes encoding the current observation, updating the hidden state and generating the LoRA adapter for the next time step, applying the adapter generated for the current time step, prefilling the model with the question and the two most recent frames from \([t-2,t]\), and decoding the first output token.

For SimpleStream (Qwen3.5-9B), the model receives only the question and two frames sampled from the most recent two-second window, \([t-2,t]\), without earlier visual history. The measured latency includes prefilling with these inputs and decoding the first output token.

For Qwen3.5-9B with full context, the measured latency includes prefilling the question together with the complete video history from the start of the stream to \(t\), sampled at 1 FPS, followed by decoding the first output token.

For VideoChat3, we follow its official streaming implementation. At each query, the model re-prefills the question together with a sliding window of the most recent 32 frames, corresponding to approximately 16 seconds at 2 FPS, before generating the streaming response. The measured latency includes prefilling of the local window video and text query and subsequent decoding.

For MOSS-VL-Realtime, we follow its official streaming implementation, in which incoming observations are continuously incorporated into a growing vision-text KV state. The measured latency includes the forward pass required to ingest the latest two seconds of video and the subsequent decoding.

\section{Additional Experiments}
\label{appendix:additional_experiments}
\subsection{Ablation Study}\label{appendix:ablation_study}

We conduct ablation studies with \method (Qwen3.5-9B) to investigate key design choices: (1) the LoRA target modules, (2) the LoRA rank, (3) the architecture of the \method, 
(4) the size of the current observation and (5) the refresh rate of the hidden state.
All ablations are conducted on a subset of an subset of the training data described in Section~\ref{section:Training_data}, rather than the final dataset used for the main experiments. 
Within each ablation, all training hyperparameters are kept fixed following Section~\ref{sec:main_exp}, except for the component being ablated, ensuring a controlled comparison. We report results on OVO-Bench and StreamingBench, together with the Episodic Memory (EPM) results from OVO-Bench.

\textit{Where should LoRA be applied?} Qwen3.5-9B adopts a hybrid attention architecture in which one self-attention layer is interleaved with every four linear-attention layers, providing a representative setting for studying LoRA placement in modern hybrid-attention VLMs. We evaluate three cumulative configurations: (1) applying LoRA only to the MLP layers shared by both self-attention and linear-attention blocks; (2) additionally applying LoRA to the \texttt{v\_proj} in self-attention layers; and (3) further applying LoRA to the \texttt{qkv\_proj} in linear-attention layers. As shown in Table~\ref{tab:lora_module}, progressively extending LoRA to additional modules consistently improves the average performance, with the best results achieved when all three module types are adapted.

\textit{What is the best LoRA rank?} The results are presented in Table~\ref{tab:lora_rank}. Among the evaluated configurations, rank 8 achieves the best overall performance, with an average score of 69.2. Rank 4 performs slightly worse, suggesting limited adaptation capacity, whereas increasing the rank to 16 causes a substantial performance drop on both benchmarks, indicating overfitting.

\begin{table*}[t]
    \centering

    \begin{minipage}[t]{0.48\textwidth}
        \vspace{0pt}
        \centering
        \caption{Ablation study on the modules to which LoRA is applied. Each row cumulatively adds LoRA to the module indicated in the first column.}
        \vspace{-0.8em}
        \label{tab:lora_module}
        \medskip
        \scalebox{0.7}{
        \begin{tabular}{lccc}
            \toprule
            Module & OVO-Bench (EPM) & StreamingBench & Avg. \\
            \midrule
            MLP                  & 63.6 (63.3) & 72.4 & 68.0 \\
            + \texttt{v\_proj}   & 63.8 (63.6) & 72.4 & 68.1 \\
            + \texttt{qkv\_proj} & 64.1 (63.2) & 72.5 & \textbf{68.3} \\
            \bottomrule
        \end{tabular}
        }
    \end{minipage}
    \hfill
    \begin{minipage}[t]{0.48\textwidth}
        \vspace{0pt}
        \centering
        \caption{Ablation study on the learned LoRA rank.}
        \vspace{-0.5em}
        \label{tab:lora_rank}
        \medskip
        \scalebox{0.8}{
        \begin{tabular}{lccc}
            \toprule
            Rank & OVO-Bench (EPM) & StreamingBench & Avg. \\
            \midrule
            4  & 65.3 (63.0) & 71.9 & 68.6 \\
            8  & 65.7 (63.6) & 72.8 & \textbf{69.2} \\
            16 & 63.4 (59.3) & 68.2 & 65.8 \\
            \bottomrule
        \end{tabular}
        }
    \end{minipage}

    \vspace{-3mm}
\end{table*}

\textit{What is the best \method module structure?}
We investigate key design choices for the \method module: (1) whether temporal modeling with a State-Space Model (SSM) and global semantic aggregation are both necessary, and (2) how global semantics should be effectively preserved.

Table~\ref{tab:ssm_and_sum} evaluates the contributions of the SSM and global semantic aggregation. The \textit{w/ sum} variant achieves stronger performance on EPM, which emphasizes information from earlier observations, whereas the \textit{w/ SSM} variant performs better on StreamingBench, where a large portion of the evaluation focuses on real-time understanding. These results suggest that the SSM is particularly effective at modeling temporally local and causal information, while global semantic aggregation better preserves information over longer temporal horizons. Combining both components balances real-time understanding and long-term memory, yielding the best overall performance.

Table~\ref{tab:weight_sum_or_gated} further compares different strategies for preserving global semantics. We consider three variants: \textit{sum}, which directly accumulates all previous compact visual representations $q_t$; \textit{weighted sum}, which weights each $q_t$ according to the cosine similarity between adjacent temporal blocks; and \textit{gated sum}, which selectively retains $q_t$ based on temporal similarity and is adopted in our final design. The \textit{gated sum} achieves the best average performance. By filtering temporally redundant representations while retaining informative observations, the gating mechanism preserves global semantics more effectively and improves overall performance.

\begin{table*}[t]
    \centering

    \begin{minipage}[t]{0.48\textwidth}
        \centering
        \caption{Ablation study on the causal relationship and global semantics.}
        \vspace{-0.5em}
        \label{tab:ssm_and_sum}
        \medskip
        \scalebox{0.7}{
        \begin{tabular}{lccc}
            \toprule
            Variant & OVO-Bench (EPM) & StreamingBench & Avg. \\
            \midrule
            w/ sum       & 62.5 (63.0)  & 70.9 & 66.7 \\
            w/ SSM       & 62.4 (60.9)  & 72.8 & 67.6 \\
            w/ SSM + sum & 64.2 (63.0) & 72.7 & \textbf{68.5} \\
            \bottomrule
        \end{tabular}
        }
    \end{minipage}
    \hfill
    \begin{minipage}[t]{0.48\textwidth}
        \centering
        \caption{Ablation study on different global semantics aggregation strategies.}
        \vspace{-0.5em}
        \label{tab:weight_sum_or_gated}
        \medskip
        \scalebox{0.7}{
        \begin{tabular}{lccc}
            \toprule
            Variant & OVO-Bench (EPM) & StreamingBench & Avg. \\
            \midrule
            sum          & 64.1 (61.3) & 72.1 & 68.1 \\
            weighted sum & 64.6 (62.0) & 72.7 & 68.6 \\
            gated sum    & 64.8 (64.0) & 72.7 & \textbf{68.8} \\
            \bottomrule
        \end{tabular}
        }
    \end{minipage}

    \vspace{-1mm}
\end{table*}

\textit{How does the size of the current observation affect performance?} Although the main experiments in Section~\ref{sec:main_exp} treat each \(C_t\) as a 2-second video clip, updating the hidden state \(h_t\) every 2 seconds, the observation window can be varied independently of \(h_t\) update frequency. To isolate the effect of the current observation window while keeping the hidden-state update interval fixed, we ablate the window sizes of 1, 2, and 4 seconds, with 2 seconds corresponding to our default setting. For the 4-second window, \(h_t\) is updated using the oldest 2-second segment evicted from the window. For the 1-second window, \method accumulates consecutive observations and updates \(h_t\) after every 2 seconds of newly observed video.
The result is presented in Table~\ref{tab:current_observation}. A larger 4-second current observation window improves real-time perception, as reflected by the higher StreamingBench score, but degrades memory performance, with EPM dropping to 60.0. In contrast, a 1-second window reduces real-time perception while yielding similar EPM performance as the 2-second window. To balance real-time perception and long-term memory, we therefore use a 2-second current observation window as the default setting.

\textit{How does the refresh rate of the hidden state affect the performance?} Although the main experiments in Section~\ref{sec:main_exp} treat each \(C_t\) as a 2-second video clip, updating the hidden state \(h_t\) every 2 seconds, the update frequency can be varied independently of the observation window. We therefore conduct an ablation with update intervals of 1, 2 (which used in the main experiments), and 4 seconds while keeping the base VLM input fixed to the most recent 2 seconds. For a 1-second interval, the hidden state is updated using the oldest 1-second segment evicted from the current 2-second observation window; for a 4-second interval, \method accumulates newly observed video and updates the hidden state only after every 4 seconds of new observations.

The results are presented in Table~\ref{tab:refresh_rate}. We observe that less frequent updates, where each update incorporates a larger temporal chunk, slightly improve memory performance at the cost of a small drop in real-time perception. We hypothesize that this behavior is related to the State-Space Model (SSM) used for hidden-state updates. As shown in Table~\ref{tab:ssm_and_sum}, the SSM primarily models short-term temporal dependencies. More frequent updates may therefore bias it toward shorter temporal ranges, improving current perception but slightly weakening longer-term memory, while less frequent updates have the opposite effect. Overall, performance remains stable across different update intervals. We therefore adopt a 2-second interval, which balances memory and real-time perception while matching the current observation window for a simpler design.

\begin{table*}[t]
    \centering

    \begin{minipage}[t]{0.48\textwidth}
        \centering
        \caption{Ablation study on the current observation size; frames are sampled at 1fps.}
        \vspace{-0.5em}
        \label{tab:current_observation}
        \medskip
        \scalebox{0.65}{
        \begin{tabular}{lccc}
            \toprule
            Variant & OVO-Bench (EPM) & StreamingBench & Avg. \\
            \midrule
            1 second (1 frame)      & 66.0 (63.2) & 71.0 & 68.5 \\
            2 seconds (2 frames)    & 66.3 (63.3) & 73.1 & 69.7 \\
            4 seconds (4 frames)     & 65.4 (\underline{60.0}) & 75.1 & \textbf{70.3} \\
            \bottomrule
        \end{tabular}
        }
    \end{minipage}
    \hfill
    \begin{minipage}[t]{0.48\textwidth}
        \centering
        \caption{Ablation study on the hidden state ($h_t$) refresh rate; frames are sampled at 1fps.}
        \vspace{-0.5em}
        \label{tab:refresh_rate}
        \medskip
        \scalebox{0.65}{
        \begin{tabular}{lccc}
            \toprule
            Variant & OVO-Bench (EPM) & StreamingBench & Avg. \\
            \midrule
            1 second (1 frame)       & 65.2 (62.8) & 73.2 & 69.2 \\
            2 seconds (2 frames)      & 66.3 (63.3)  & 73.1 & \textbf{69.7} \\
            4 seconds (4 frames)      & 66.5 (63.6) & 72.3 & 68.4 \\
            \bottomrule
        \end{tabular}
        }
    \end{minipage}
    \vspace{-3mm}
\end{table*}

\subsection{Additional Analysis} \label{appendix:additional_analysis}

\textbf{Latency Breakdown.}
We further analyze the latency of \method (Qwen3.5-9B) reported in Figure~\ref{fig:speed}, with the detailed breakdown presented in Table~\ref{tab:c2l-latency-breakdown}. Encoding the new frames and updating the hidden state $h_t$, generating the LoRA tensors $A_{t+1}$ and $B_{t+1}$, and combining them to construct $\Delta\theta_{t+1}$ account for 7.9\%, 2.1\%, and 3.7\% of the total latency, respectively. Importantly, these operations compute the adapter for the next temporal block and can therefore be executed in parallel with applying $\Delta\theta_t$ and performing base-VLM prefilling and decoding for the current block. With such pipelining, only the application of $\Delta\theta_t$ and the wrapper overhead remain on the critical inference path, adding 1.75~ms to the 130.07~ms base-model inference latency. This corresponds to approximately 1.3\% of the effective end-to-end latency, demonstrating that \method introduces only marginal latency overhead in practical streaming inference.

\begin{table}[t]
\centering

\caption{Breakdown of the \method (Qwen3.5-9B) latency. Percentages are relative to the total latency of 152.89 ms.}
\vspace{-0.5em}
\label{tab:c2l-latency-breakdown}
\medskip
\scalebox{0.9}{
\begin{tabular}{lrr}
\toprule
\textbf{Component} & \textbf{Latency (ms)} & \textbf{Percentage} \\
\midrule
\multicolumn{3}{l}{\itshape\small\textbf{Could be paralleled with the base VLMs prefilling and decoding}} \\
Encode new frames and update state $h_t$ & 12.11 & 7.91\% \\
Generate LoRA tensors $A_{t+1}$ and $B_{t+1}$        &  3.27 & 2.14\% \\
Combine LoRAs and generate $\Delta\theta_{t+1}$                     &  5.69 & 3.72\% \\
\midrule
\multicolumn{3}{l}{\itshape\small\textbf{Can not be paralleled with the base VLMs prefilling and decoding}} \\
Apply LoRA $\Delta\theta_{t}$                         &  1.13 & 0.74\% \\
Wrapper overhead                   &  0.62 & 0.41\% \\
Base-model prefill + decoding first tokens                 & 130.07 & 85.07\% \\
\midrule
\textbf{Total}                     & \textbf{152.89} & \textbf{100.00\%} \\
\bottomrule
\end{tabular}
}

\end{table}

\begin{figure*}[t]
    \centering
    \includegraphics[width=0.85\linewidth]{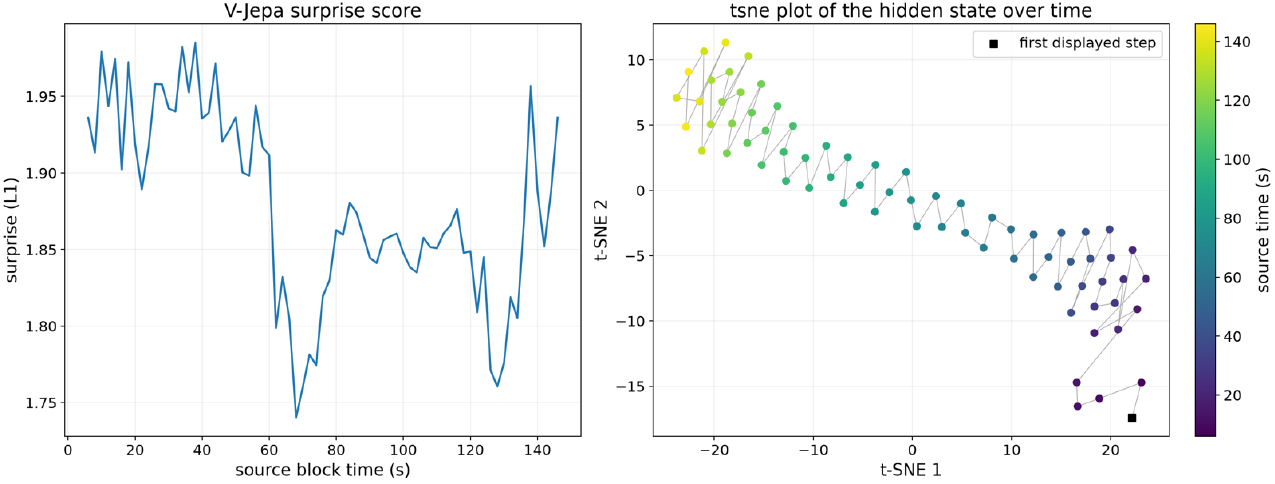}
    \vspace{-0.5em}
    \caption{The V-JEPA-2 surprise score versus the t-SNE map of the LoRA hidden state.}
    \label{fig:lora_evolve}
    \vspace{-0.5em}
\end{figure*}

\textbf{LoRA Hidden State Evolution over the History.} Figure~\ref{fig:lora_evolve} visualizes how the internal LoRA state of \method evolves with changes in video content. We compute a V-JEPA-2~\citep{assran2025v} surprise score on a YouCook2~\citep{ZhXuCoAAAI18} video by masking the features of the last temporal block and measuring the distance between its predicted representation and the representation of the subsequent block. In parallel, we track the 512-dimensional LoRA hidden state of \method (Qwen3.5-9B), defined as $\mathrm{Norm}(o_t)+\alpha\mathrm{Norm}(s_t)$. We average the hidden states used to generate layer-specific LoRA adapters and project the resulting representation into two dimensions using t-SNE.

The LoRA state changes more rapidly when the V-JEPA-2 surprise score is high (before 60s and after 130s), while evolving more smoothly during the lower-surprise interval between 60s and 130s. This correspondence suggests that the LoRA state adapts dynamically to changes in the underlying video events.

\textbf{Effect of Memory on Real-Time Perception.}
We investigate whether long-term historical information compressed by \method adversely affects real-time perception on the real-time splits of OVO-Bench and StreamingBench.

\begin{figure*}[t]
    \centering
    \includegraphics[width=0.9\linewidth]{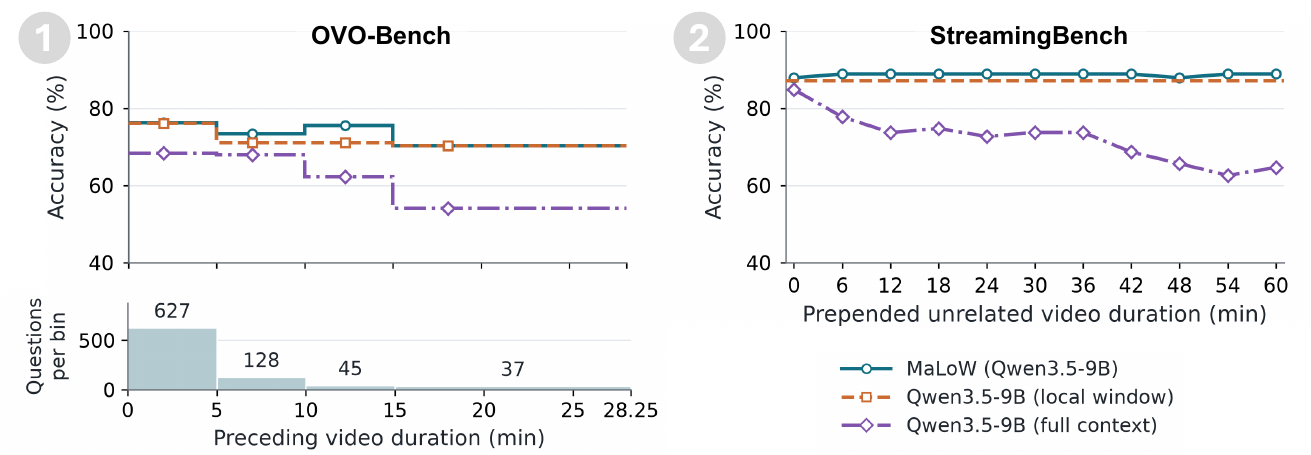}
    \vspace{-0.5em}
    \caption{Effect of Memory on Real-Time Perception. Markers on subplot 1 show bin medians. We evaluate the accuracy of \method on real-time questions across varying history lengths. \method maintains stable real-time perception despite accumulating long-term memory.}
    \label{fig:appendix_memory_compete_with_perception}
    \vspace{-0.5em}
\end{figure*}

For OVO-Bench, where video prefixes extend up to 28 minutes, we group real-time questions by their preceding video duration into four intervals: 0–5, 5–10, 10–15, and over 15 minutes. As shown in Figure~\ref{fig:appendix_memory_compete_with_perception}-1, \method (Qwen3.5-9B) maintains comparable accuracy to the local-window baseline (Qwen3.5-9B, last 2 seconds) across different prefix ranges, suggesting that accumulated memory does not compromise real-time perception.

For StreamingBench, where the original video prefixes are relatively short, we randomly sample 100 samples and construct longer streams by prepending randomly sampled videos, extending the historical context from 5 to 60 minutes. As shown in Figure~\ref{fig:appendix_memory_compete_with_perception}-2, consistent with the observations on OVO-Bench, \method (Qwen3.5-9B) maintains stable real-time perception across varying history lengths. 


These observations are consistent with our design choice to regularize the norm of the generated LoRA weights. This regularization constrains the magnitude of memory-induced weight updates, preventing them from dominating the pretrained model weights and disrupting its real-time perception capabilities.

\begin{figure*}[t]
    \centering
    \includegraphics[width=0.99\linewidth]{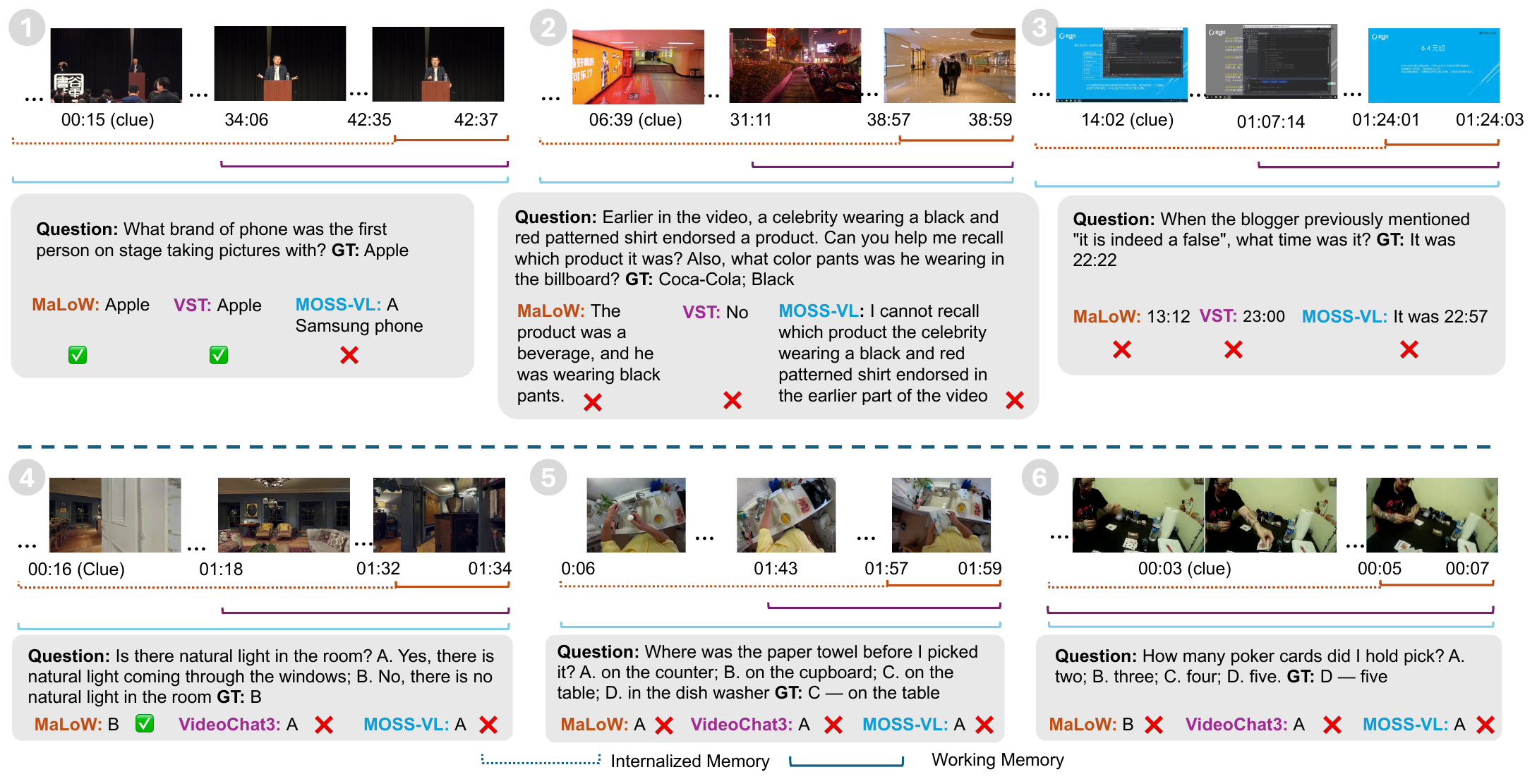}
    \vspace{-0.5em}
    \caption{The additional qualitative results.}
    \label{fig:appendix_additional_qualitative_result}
    \vspace{-0.5em}
\end{figure*}

\textbf{Additional Qualitative Results.}
We provide additional qualitative results in Figure~\ref{fig:appendix_additional_qualitative_result}. Samples 1, 2, and 4 further demonstrate that \method (Qwen3.5-9B) can retain long-term historical context through its internalization mechanism. In some cases, however, the retained memory can be vague, as humans do. For example, in Sample 2, which clue appeared 32 minutes before the question timing, \method appears to retain a coarse memory of Coca-Cola and answers ``beverage'' rather than the specific object.

Sample 5 illustrates a failure case in which both \method and the baselines produce incorrect answers, partly due to an inaccurate ground-truth annotation. Sample 3 requires not only remembering the relevant video timestamp, but also an agentic procedure to revisit the corresponding frames and localize the clock displayed in the bottom-right corner. Since \method is not trained with agentic tool-use data, extending it to such agentic settings is left for future work. Finally, Sample 6 requires strong visual perception to recognize that four cards are initially present on the table, followed by temporal reasoning to determine that one additional card is added, yielding five cards in total. We expect such cases to benefit from future base VLMs with stronger fine-grained visual perception.

\end{document}